\documentclass[acmsmall]{acmart}

\usepackage{color}

\def\cll{\textcolor{black}}

\def\ah{\textcolor{magenta}}

\newcommand{\sign}{\text{sign}}

\usepackage{graphicx}
\usepackage{multirow}
\usepackage{ulem}

\usepackage{printlen}
\usepackage{subfigure}
\usepackage{soul} % added by ahoud in order to use Strikethrough text

\def\BibTeX{{\rm B\kern-.05em{\sc i\kern-.025em b}\kern-.08emT\kern-.1667em\lower.7ex\hbox{E}\kern-.125emX}}
    
\begin{document}

%
% The "title" command has an optional parameter, allowing the author to define a "short title" to be used in page headers.
\title{Adversarial Attacks on Deep Learning Models in Natural Language Processing: A Survey}

%
% The "author" command and its associated commands are used to define the authors and their affiliations.
% Of note is the shared affiliation of the first two authors, and the "authornote" and "authornotemark" commands
% used to denote shared contribution to the research.
\author{Wei Emma Zhang}
\email{w.zhang@mq.edu.au}
\orcid{1234-5678-9012}
\author{Quan Z. Sheng}
\email{michael.sheng@mq.edu.au}
\orcid{1234-5678-9012}
\author{Ahoud Alhazmi}
\email{ahoud.alhazmi@hdr.mq.edu.au}
\affiliation{%
  \institution{Macquarie University}
  \city{Sydney}
  \country{Australia}
  \postcode{NSW 2109}
}

% \author{Ahoud Abdulrahmn F Alhazmi}
% \email{ahoud.alhazmi@hdr.mq.edu.au}
% \affiliation{%
%   \institution{Macquarie University}
%   \city{Sydney}
%   \country{Australia}
%   \postcode{NSW 2109}
% }

\author{Chenliang Li}
\affiliation{%
  \institution{Wuhan University}
 % \streetaddress{1 Th{\o}rv{\"a}ld Circle}
  \city{Wuhan}
  \country{China}}
\email{cllee@whu.edu.cn}

%
% By default, the full list of authors will be used in the page headers. Often, this list is too long, and will overlap
% other information printed in the page headers. This command allows the author to define a more concise list
% of authors' names for this purpose.
\renewcommand{\shortauthors}{Zhang et al.}

%
% The abstract is a short summary of the work to be presented in the article.
\begin{abstract}
With  the  development  of  high  computational  devices, deep  neural  networks (DNNs),  in  recent  years,  have  gained  significant  popularity  in  many  Artificial  Intelligence  (AI)  applications.  
However, previous efforts have shown that DNNs were vulnerable to  strategically modified samples, named \textit{adversarial examples}. These samples are  generated with some imperceptible perturbations, but can fool the DNNs to give false predictions. 
Inspired by the popularity of generating adversarial examples for image DNNs, research efforts on attacking DNNs for textual applications emerges in recent years. However,   existing perturbation methods for images cannot be directly applied to texts as  text data is discrete \cll{in nature}. 
In this article, we review research works that address this difference and generate textual adversarial examples on DNNs. We collect, select, summarize, discuss and analyze these works in a comprehensive way and cover all the related information to make the article self-contained. Finally, drawing on the reviewed literature, we provide further discussions and suggestions on this topic. 
\end{abstract}



%
% Keywords. The author(s) should pick words that accurately describe the work being
% presented. Separate the keywords with commas.

%
% A "teaser" image appears between the author and affiliation information and the body 
% of the document, and typically spans the page. 
%%  \includegraphics[width=\textwidth]{sampleteaser}
%%  \caption{Seattle Mariners at Spring Training, 2010.}
%%  \Description{Enjoying the baseball game from the third-base seats. Ichiro Suzuki preparing to bat.}

%
% This command processes the author and affiliation and title information and builds
% the first part of the formatted document.
\maketitle

\section{Introduction}

Deep neural networks (DNNs) are large neural networks \cll{whose architecture is organized as a series of} layers of neurons, \cll{each of which serves as} the  individual  computing  units. Neurons are connected by links with different weights and biases and transmit the results of its activation function on its inputs to \cll{the neurons of the next layer}. Deep neural networks try to mimic the biological neural networks of human brains to learn and build knowledge from examples. Thus they are shown the strengths in dealing with complicated tasks that are not easily to be modelled as linear or non-linear problems.  Further more, \cll{empowered by continuous real-valued vector representations (i.e.,  embeddings)} they are good at handling data with various modalities, e.g., image, text, video and audio. 

With the development of high computational devices, deep neural networks, in recent years have  gained significant popularity in many Artificial Intelligence (AI) communities such as Computer Vision \cite{nips/KrizhevskySH12,corr/SimonyanZ14a}, Natural Language Processing \cite{icml/KumarIOIBGZPS16,conll/BowmanVVDJB16}, Web Mining \cite{taslp/PalangiDSGHCSW16,corr/YangYHGD14a} and Game theory \cite{nips/SchuurmansZ16}. 
% nn blackbox
However, the interpretability of deep neural networks is still unsatisfactory as they work as black boxes, which means it is difficult to get intuitions from what each neuron exactly has learned. One of the problems of the poor interpretability is evaluating the robustness of deep neural networks. 
%\att{add some history of attack on non-DNN.}
% 
% NN not robust 
In recent years, research works \cite{iclr14/Szegedy,iclr15/goodfellow} used small unperceivable perturbations to evaluate the robustness of deep neural  networks and found that they are not robust to these perturbations. 
% example works
Szegedy et al. \cite{iclr14/Szegedy} first evaluated the  state-of-the-art deep neural networks used for image classification with small generated perturbations on the input images. They found that the image classifier were fooled with high probability, but human judgment is not affected. The perturbed image pixels were named \textit{adversarial examples} and this notation is later used to denote all kinds of perturbed samples in a general manner. As the generation of adversarial examples is costly and impractical in \cite{iclr14/Szegedy}, Goodfellow et al. \cite{iclr15/goodfellow} proposed a fast generation method which popularized this research topic (Section \ref{sec:sub_cv} provides further discussion on these works). Followed their works, many research efforts have been made and the purposes of these works can be summarized as: 
i) evaluating the deep neural networks by fooling them with unperceivable perturbations;
ii) intentionally changing the output of the deep neural networks;
and iii) detecting the oversensitivity and over-stability points of the deep neural networks and finding solutions to defense the attack.

%to text
Jia and Liang \cite{emnlp/JiaL17} are the first to consider adversarial example generation (or \textit{adversarial attack}, we will use these two expressions interchangeably hereafter) on \cll{deep neural networks for text-based tasks} (namely \textit{textual deep neural networks}). Their work quickly gained research attention in Natural Language Processing (NLP) community. 
However, due to \cll{intrinsic} differences between images and textual data, the adversarial attack methods on images cannot be directly applied to the latter one. 
First of all, image data (e.g., pixel values) is continuous, but textual data is discrete. \cll{Conventionally}, we vectorize the texts before inputting them into the deep neural networks. Traditional vectoring methods include leveraging term frequency and inverse document frequency, and one-hot representation (details in Section \ref{sec:sub_vectorizing}). When applying gradient-based adversarial attacks adopted from images on these representations, the generated adversarial examples are invalid characters or word sequences \cite{zhao2017generating}. \cll{One solution is to use word embeddings as the input of deep neural networks}. \cll{However, this will also generate words that can not be matched with any words in the word embedding space \cite{gong2018adversarial}}. 
Secondly, the perturbation of images are small change of pixel values that are hard to be perceived by human eyes, thus humans can correctly classify the images, showing the poor robustness of deep neural models. But for adversarial attack on texts, small perturbations are easily perceptible. For example, replacement of characters or words would generate invalid words or syntactically-incorrect sentences. Further, it would alter the semantics of the sentence drastically. 
Therefore, the perturbations are easily to be perceived--in this case, even human being cannot provide correct predictions. 

To address the aforementioned differences and challenges, many attacking methods are proposed since the pioneer work of Jia and Liang \cite{emnlp/JiaL17}.
Despite the popularity of the topic in NLP community, there is  no comprehensive review paper that collect and summarize the  efforts in this research direction. There is a need for this kind of work that helps successive researchers and practitioners to have an overview of these methods.

\vspace{2mm}
\noindent 
\textbf{Related surveys and the differences to this survey. }  
In \cite{journals/ml/BarrenoNJT10}, the authors presented comprehensive review on different classes of attacks and defenses against machine learning systems. Specifically, they proposed a taxonomy for identifying and analyzing these attacks and applied the attacks on a machine learning based application, i.e., a statistical spam filter, to illustrate the effectiveness of the attack and defense. This work targeted machine learning algorithms rather than neural models. 
Inspired by \cite{journals/ml/BarrenoNJT10}, the authors in \cite{corr/GilmerAGAD18} reviewed the defences of adversarial attack in the security point of view. The work is not limited to machine learning algorithms or neural models, but a generic report about adversarial defenses on security related applications. The authors found that existing security related defense works lack of clear motivations and explanations on how the attacks are related to the real security problems and how the attack and defense are meaningfully evaluated. Thus they established a taxonomy of motivations, constraints, and abilities for more plausible adversaries. %They provided a series of recommendations for future works. 
%This work focuses on security applications and does not distinguish continuous or discrete data.
 %
\cite{pr/BiggioR18} provides a thorough overview of the evolution of the adversarial attack research over the last ten years, and focuses on the research works from computer vision and cyber security. The paper covers the works from pioneering non-deep leaning algorithms to recent deep learning algorithms. It is also from the security point of view to provide detailed analysis on the effect of the attacks and defenses. 
The authors of \cite{access/LiuLZCYL18} reviewed the same problem in a data-driven perspective. They analyzed the attacks and defenses according to the learning phases, i.e., the training phase and test phase. 

Unlike previous works that discuss generally on the attack methods on  machine learning algorithms, \cite{corr/abs-1712-07107} focuses on the adversarial examples on deep learning models. It reviews current research efforts on attacking various deep neural networks in different applications. The defense methods are also extensively surveyed. 
However, they mainly discussed adversarial examples for image classification and object recognition tasks. 
The work in \cite{access/AkhtarM18} provides a comprehensive review on the adversarial attacks on deep learning models used in computer vision tasks. It is an application-driven survey that groups the attack methods according to the sub-tasks under computer vision area. The article also comprehensively reports the works on \cll{the defense side}, the methods of which are mainly grouped into three categories.  

All the mentioned works either target general overview of the attacks and defenses on machine learning models or focus on specific domains such as  computer vision and cyber security. Our work differs with them that we specifically focus on the attacks and defenses on textual deep learning models. Furthermore, we provide a comprehensive review that covers information from different aspects to make this survey self-contained.

\vspace{2mm}
\noindent 
\textbf{Papers selection.} 
The papers we reviewd in this article are high quality papers selected from top NLP and AI conferences, including ACL\footnote{Annual Meeting of the Association for Computational Linguistics}, COLING\footnote{International Conference on Computational Linguistics}, NAACL\footnote{Annual Conference of the North American Chapter of the Association for Computational Linguistics}, EMNLP\footnote{Empirical Methods in Natural Language Processing}, ICLR\footnote{International Conference on Learning Representations}, AAAI\footnote{AAAI Conference on Artificial Intelligence} and IJCAI\footnote{International Joint Conference on Artificial Intelligence}. 
Other than accepted papers in aforementioned conferences, we also consider good papers in e-Print archive\footnote{arXiv.org}, as it reflects the latest research outputs. We select papers from archive with three metrics: paper quality, method novelty and the number of citations (optional\footnote{As the research topic emerges from 2017, we relax the citation number to over five if it is published more than one year. If the paper has less than five citations, but is very recent and satisfies the other two metrics, we also \cll{include it in this paper}.}).

\vspace{2mm}
\noindent 
\textbf{Contributions of this survey.} 
The aim of this survey is to provide a comprehensive review on the  research efforts on generating adversarial examples on textual deep neural networks. It is motivated by the drastically increasing attentions on this topic. This survey will serve  researchers and practitioners who are interested in attacking textual deep neural models. More broadly, it can serve as a reference on how deep learning is applied in NLP community. We expect that the readers have some basic knowledge of the deep neural networks architectures, which are not the focus in this article. 
To summarize, the key contributions of this survey are:
\begin{itemize}
\item We conduct a comprehensive review for adversarial attacks on textual deep neural models and propose different classification schemes to organize the reviewed literature; this is the first work of this kind;  
\item We provide all related information to make the survey self-contained and thus it is easy for readers who have limited NLP knowledge to understand;  
\item  We discuss some open issues, and identify some possible research directions in this research field aims to build more robust textual deep learning models with the help of adversarial examples. 
\end{itemize}

The remainder of this paper is organized as follows: We introduce the preliminaries for adversarial attacks on deep learning models in Section \ref{sec:overview} including the taxonomy of adversarial attacks and deep learning models used in NLP. In Section \ref{sec:fromto}, we address the difference of attacking image data and textual data and briefly reviewed exemplary works for attacking image DNN that inspired their follow-ups in NLP.  
Section \ref{sec:attack} first presents our 
classification on the literature and then gives a detailed introduction to the state of the art.
We discuss the defense strategies in Section \ref{sec:defense} and point out the open issues in Section \ref{sec:future}. Finally, the article is concluded in Section \ref{sec:conclude}.

\section{Overview of Adversarial Attacks and Deep Learning Techniques in Natural Language Processing}
\label{sec:overview}

Before we dive into the details of this survey, we start with an introduction to the general taxonomy of adversarial attack on deep learning models. We also introduce the deep learning techniques and their applications in natural language processing.

\subsection{Adversarial Attacks on Deep Learning Models: The General Taxonomy }
In this section, we provide the definitions of adversarial attacks and introduce different aspects of the attacks, followed by the measurement of perturbations and the evaluation metrics of the effectiveness of the attacks in a general manner that applies to any data modality. 

\subsubsection{Definitions}

\begin{itemize}
    \item Deep Neural Network (DNN).  
    A deep neural network (we use DNN and deep learning model interchangeably hereafter) can be simply presented as a nonlinear function $f_\theta: \mathbf{X}\rightarrow\mathbf{Y}$, where $\mathbf{X}$ is the input features/attributes, $\mathbf{Y}$ is the output predictions that can be a discrete set of classes or a sequence of objects. $\mathbf{\theta}$ represents the DNN parameters and are learned via gradient-based back-propagation during the model training. Best parameters would be obtained by minimizing the the gap between the model's prediction $f_\mathbf{\theta}(\mathbf{X})$ and the correct label $\mathbf{Y}$, where the gap is measured by loss function $J(f_\mathbf{\theta}(\mathbf{X}),\mathbf{Y})$. 

   \item Perturbations. Perturbations are intently created small noises that to be added to the original input data examples in test stage,  aiming to fool the deep learning models. 
   
    \item Adversarial Examples. 
    An adversarial example $\mathbf{x}'$ is an example created via worst-case perturbation of the input to a deep learning model. An ideal DNN would still assign correct class $\mathbf{y}$ (in the case of classification task) to $\mathbf{x}'$, while a victim DNN would have high confidence on wrong prediction of $\mathbf{x}'$. 
    $\mathbf{x}'$ can be formalized as:

    % \begin{eqnarray}
    % \label{eq:adv}
    % \mathbf{x}'=\mathbf{x}+\mathbf{\eta}, d(\mathbf{x},\mathbf{x}')<\epsilon, \mathbf{x} \in \mathbf{X} \\
    % f(\mathbf{x}) \neq f(\mathbf{x}') \textrm{ if untargeted} \nonumber \\
    % \textrm{or } f(\mathbf{x}')=\mathbf{y}' \textrm{ if targeted} \nonumber
    % \end{eqnarray}
    
%   \vspace{-2mm} 
%   \begin{eqnarray}
%     \label{eq:adv}
%     &\mathbf{x}'=\mathbf{x}+\mathbf{\eta}, d(\mathbf{x},\mathbf{x}')<\epsilon, \mathbf{x} \in \mathbf{X} \\
%     &f(\mathbf{x}) \neq f(\mathbf{x}')  \nonumber \\
%     &\textrm{or } f(\mathbf{x}')=\mathbf{y}' \nonumber
%     \end{eqnarray}

   \vspace{-2mm} 
   \begin{eqnarray}
    \label{eq:adv}
    &\mathbf{x}'=\mathbf{x}+\mathbf{\eta},  f(\mathbf{x})=\mathbf{y}, \mathbf{x} \in \mathbf{X} \\
    &f(\mathbf{x}') \neq \mathbf{y}  \nonumber \\
    &\textrm{or } f(\mathbf{x}')=\mathbf{y}', \mathbf{y}' \neq \mathbf{y} \nonumber
    \end{eqnarray}

    \noindent where $\eta$ is the worst-case perturbation. The goal of the adversarial attack can be deviating the label to incorrect one ($f(\mathbf{x}') \neq \mathbf{y}$)  or  specified one ($f(\mathbf{x}')=\mathbf{y}'$). 
  
 \end{itemize}

\subsubsection{Threat Model}
\label{sec:threat}
We adopt the definition of \textit{Threat Model} for attacking DNN from \cite{corr/abs-1712-07107}. In this section, we discuss several aspects of the threat model. 

\begin{itemize}
    \item Model Knowledge.  The adversarial examples can be generated using black-box or white-box strategies in terms of the knowledge of the attacked DNN.
     Black-box attack is performed when the architectures, parameters, loss function, activation functions and training data of the DNN are not accessible. Adversarial examples are generated by directly accessing the test dataset, or by querying the DNN and checking the output change. On the contrary, white-box attack is based on the knowledge of certain aforementioned information of DNN.  
    \item Target. The generated adversarial examples can change the output prediction to be incorrect or to specific result as shown in Eq. (\ref{eq:adv}). Compared to the un-targeted attack  %(\ah{(No-Need space)} 
    ($f(\mathbf{x}') \neq \mathbf{y}$),
    targeted attack ($f(\mathbf{x}')=\mathbf{y}'$) is more strict as it not only changes the prediction, but also enforces constraint on the output to generate specified prediction. For binary tasks, e.g., binary classification, un-targeted attack equals to the targeted attack. 
    
    \item Granularity. The attack granularity refers to the level of data on which the adversarial examples are generated from. %\ah 
    For example, it is usually the image pixels for image data. Regarding the textual data, it could be character, word, and sentence-level embedding. Section \ref{sec:sub_vectorizing} will give further introduction on attack granularity for textual DNN. 

    \item Motivation. Generating adversarial examples is motivated by two goals: attack and defense. %\ah{The}\str{A}\ah{a}ttack 
    The attack aims to examine the robustness of the target DNN, while the defense takes a step further utilizing generated adversarial examples to robustify the target DNN. Section \ref{sec:defense} will give more details. 

\end{itemize}

\subsubsection{Measurement}
\label{sec:subsub_measure}
Two groups of measurements are required in the adversarial attack for i) controlling the perturbations and ii) evaluating the effectiveness of the attack, respectively.

\begin{itemize}
    \item Perturbation Constraint. As aforementioned, the perturbation $\eta$ should not change the true class label of the input - that is, an ideal DNN classifier, if we take classification as example, will provide the same prediction on the adversarial example to the original example. $\eta$ cannot be too small as well, to avoid ending up with no affect on target DNNs. Ideally, effective perturbation is the maximum value in a constrained range.  
    \cite{iclr14/Szegedy} firstly put a constraint that $(\mathbf{x}+\mathbf{\eta}) \in [0, 1]^n$ for image adversarial examples, ensuring the adversarial example has the same range of pixel values as the original data \cite{warde201611}. \cite{iclr15/goodfellow} simplifies the solution and use max norm to constrain $\mathbf{\eta}$: $||\mathbf{\eta}||_\infty \leq \epsilon$. This was inspired by the intuitive observation that  a perturbation which does not change any specific pixel by more than some amount $\epsilon$ cannot change the output class  \cite{warde201611}. Using max-norm is sufficient enough for image classification/object recognition tasks. Later on, other norms, e.g., $L_2$ and $L_0$,  were used to control the perturbation in attacking DNN in computer vision. Constraining $\mathbf{\eta}$ for textual adversarial attack is somehow different. Section \ref{sec:sub_vectorizing} will  give  more details.
    
    \item Attack Evaluation. Adversarial attacks are designed to degrade the performance of DNNs. Therefore, evaluating the effectiveness of the  attack is based on the performance metrics of different tasks. For example, classification tasks has metrics such as accuracy, F1 score and AUC score. We leave the metrics for different NLP as  out-of-scope content in this article and suggest readers refer to specific tasks for information.   
    
\end{itemize}

\subsection{Deep Learning in NLP} 
\label{sec:sub_texNN}

Neural networks have been gaining increasing popularity in NLP community in recent years and various DNN models have been adopted in different NLP tasks. 
Apart from the feed forward neural networks and Convolutional Neural Networks (CNN),  Recurrent/Recursive Neural Networks (RNN) and their variants are the most common neural networks used in NLP, because of their natural ability of handling sequences.  
In recent years, two important breakthroughs in deep learning are brought into NLP. They are sequence-to-sequence learning \cite{nips/SutskeverVL14} and attention modeling \cite{corr/BahdanauCB14}. Reinforcement learning  and generative models are also gained much popularity \cite{cim/YoungHPC18}. 
%\ah{(do we need to put reference here?)}.
%
%However, as DNN, CNN and RNN are the most mature neural models in NLP, the adversarial examples are only examined on these models so far. We will provide further discussion in Section \ref{sec:future}.  
%
In this section, we will briefly overview the DNN architectures and techniques applied in NLP that are closely related to this survey. 
We suggest readers refer to detailed reviews of neural networks in NLP in \cite{corr/abs-1807-10854,cim/YoungHPC18}.%,link/standford_CS224n

\subsubsection{Feed Forward Networks}
Feed-forward network, in particular multi-layer perceptrons (MLP), is the simplest neural network.
It has several forward layers and each node in a layer connects to each node in the following layer, making the network fully connected. 
MLP utilizes nonlinear activation function to  distinguish data that is not linearly separable. 
MLP works with fixed-sized inputs and do not record the order of the elements. Thus it is mostly used in the tasks that can be formed as supervised learning problems. In NLP, it can be used in any application.  % such as text classification  machine translation and speech recognition.  % \ah{(is it NLP task?)},. 
The major drawback of feed forward networks in NLP is that it cannot handle well the text sequences in which the word order matters. 

As the feed forward network is easy to implement, there are various implementations and no standard benchmark architecture worth examining. To evaluate the robustness of feed forward network in NLP, adversarial examples are often generated for specific architectures in real applications. For example, authors of \cite{grosse2016adversarial,ESORICS17/GrosseAdvMalware,huang2018adversarial} worked on the specified malware detection models. % and \cite{acl/WallaceB18} was on question answering. 

\subsubsection{Convolutional Neural Network (CNN) }
Convolutional Neural Network contains convolutional layers and pooling (down-sampling) layers and final fully-connected layer. Activation functions are used to connect the down-sampled layer to the next convolutional layer or fully-connected layer. CNN allows arbitrarily-sized inputs. 
Convolutional layer uses convolution operation to extract meaningful local patterns of input. Pooling layer reduces the parameters %\ah{
and computation in the network and it allows the network to be deeper and less-overfitting. 
Overall, CNN identifies local predictors and combines them together to generate a fixed-sized vector for the inputs, which contains the most  %\ah{or important}
or important informative aspects for the application task. In addition, it is order-sensitive. Therefore, it excels in computer vision tasks and later is widely adopted in NLP applications. 

Yoon Kim \cite{emnlp/Kim14} adopted CNN for sentence classification. He used Word2Vec to represent words as input. Then the convolutional operation is limited to the direction of word sequence, rather than the word embeddings. Multiple filters in pooling layers deal with the  variable length of sentences. The model demonstrated excellent performances on several benchmark datasets against multiple state-of-the-art works. 
This work became a benchmark work of adopting CNN in NLP applications. 
% "In Text Understanding from Scratch, Xiang Zhang and Yann LeCun \cite{\cite{corr/ZhangL15} }, demonstrate that CNNs can achieve outstanding performance without the knowledge of words, phrases, sentences and any other syntactic or semantic structures with regards to a human language"
%
Zhang et al. \cite{nips/ZhangZL15} presented CNN for text classification at character level. They used one-hot representation in alphabet for each of the character. To control the generalization error of the proposed CNN, they additionally performed data augmentation by replacing words and phrases with their synonyms. %Evaluations on eight datasets showed that character-level CNN works better, compared to word-level CNN in  \cite{emnlp/Kim14} , for less curated user-generated texts. 
These two representative textual CNNs are evaluated via adversarial examples in many applications \cite{liang2017deep,gao2018black,acl/EbrahimiRLD18,belinkov2017synthetic,coling/EbrahimiLD18}. %yang2018greedy

\subsubsection{Recurrent Neural Networks/ Recursive Neural Networks}
Recurrent Neural Networks are neural models adapted from feed-forward neural networks for learning mappings between sequential inputs and outputs  \cite{rumelhart1986learning}. 
RNNs allows data with arbitrary length and it introduces cycles in their computational graph to model efficiently the influence of time  \cite{goodfellow2016deep}. The model does not suffer from statistical estimation problems stemming from data sparsity and thus leads to impressive performance in dealing with sequential data \cite{synthesis/2017Goldberg}.  %
Recursive neural networks \cite{goller1996learning} extends recurrent neural networks from sequences to tree, which respects the hierarchy of the language. 
% Bi-RNN
In some situations, backwards dependencies exist, which is in need for the backward analysis. Bi-directional RNN thus was proposed for looking at sentences in both directions, forwards and backwards, using two parallel RNN networks, and combining their outputs.  %  
%
% !!The presence of cyclical computations potentially
% presents challenges to the applicability of existing adversarial
% sample algorithms based on model differentiation, as cycles
% prevent computing gradients directly by applying the chain
% rule. 
% !!The presence of cyclical computations potentially
% presents challenges to the applicability of existing adversarial
% sample algorithms based on model differentiation, as cycles
% prevent computing gradients directly by applying the chain
% rule. 
%
%
Bengio et al. \cite{jmlr/BengioDVJ03} is one of the first to apply RNN in NLP. Specifically, they utilized RNN in language model, where the probability of a sequence of words is computed in an recurrent manner. The input to RNN is the feature vectors for all the preceding words, and the output is the conditional probability distribution over the output vocabulary. Since RNN is a natural choice to model various kinds of sequential data, it has been applied to many NLP tasks. Hence RNN has drawn great interest for adversarial attack \cite{papernot2016crafting}. 

RNN has many variants, among which Long Short-Term Memory (LSTM) network \cite{neco/HochreiterS97} gains the most popularity.  
LSTM is a specific RNN that was designed to capture the long-term dependencies. In LSTM, the hidden state are computed through combination of three \textit{gates}, i.e., \textit{input gate}, \textit{forget gate} and \textit{output gate}, that control information flow drawing on the logistic function. 
LSTM networks have subsequently proved to be more effective than conventional RNNs \cite{icassp/GravesMH13}. 
GRUs is a simplified version of LSTM that it only consists two gates, thus it is more efficient in terms of training and prediction. 
Some popular LSTM variants are proposed to solve various NLP tasks \cite{neco/HochreiterS97,acl/TaiSM15,corr/WangJ16a,acl/ChenZLWJI17,corr/WuSCLNMKCGMKSJL16,iclr/RocktaschelGHKB16,acl/ChenZLWJI17}. These representative works received the interests of evaluation with adversarial examples recently \cite{gao2018black,kdd/SunTYWZ18,sato2018interpretable,papernot2016crafting,naacl/IyyerWGZ18,emnlp/JiaL17,zhao2017generating,conll/Minervini018,iclr/RocktaschelGHKB16}. % yang2018greedy

% \cite{gao2018black,kdd/SunTYWZ18,sato2018interpretable} attacked self-implemented valina LSTMs. 
% \cite{papernot2016crafting,yang2018greedy,naacl/IyyerWGZ18} attacked LSTM proposed in  \cite{neco/HochreiterS97,acl/TaiSM15} respectively.
% \cite{emnlp/JiaL17} attacked Match-LSTM \cite{corr/WangJ16a}.  \cite{zhao2017generating} attacked TreeLSTM \cite{acl/ChenZLWJI17} and Google Translation system \cite{corr/WuSCLNMKCGMKSJL16}. 
% \cite{conll/Minervini018} attacked  conditional Bidirectional  LSTM (cBiLSTM) \cite{iclr/RocktaschelGHKB16} and Enhanced LSTM model (ESIM) \cite{acl/ChenZLWJI17}. 

\subsubsection{Sequence-to-Sequence Learning (Seq2Seq) Models}
Sequence-to-sequence learning (Seq2Seq) \cite{nips/SutskeverVL14} is one of the important breakthroughs in deep learning and is now widely used for NLP applications. Seq2Seq model has the superior capacity to generate another sequence information for a given sequence information with an encoder-decoder architecture~\cite{springer/2018DengL}. 
Usually, a Seq2Seq model consists of two recurrent neural networks: an encoder that processes the input and compresses it into a vector representation, a decoder that  predicts the output. 
Latent Variable Hierarchical Recurrent Encoder-Decoder (VHRED) model~\cite{aaai/SerbanSBCP16} is a recently popular Seq2Seq model that generate sequences leveraging the complex dependencies between subsequences. 
\cite{emnlp/ChoMGBBSB14} is one of the first neural machine translation (NMT) model that adopt the Seq2Seq model. 
OpenNMT \cite{acl/KleinKDSR17}, a Seq2Seq NMT model proposed recently, becomes one of the benchmark works in NMT. 
As they are adopted and applied widely, attack works also emerge \cite{conll/NiuB18,acl/EbrahimiRLD18,cheng2018seq2sick,acl/SinghGR18}.  % corr/HeC2018,,acl/WallaceB18

% \cite{conll/NiuB18} attacked Latent Variable Hierarchical Recurrent Encoder-
% Decoder (VHRED) model\cite{aaai/SerbanSBCP16} via attention mechanism \cite{iclr/BahdanauCB15}. Although this work attacked attention-based model, their attack strateies are not designed for attention model, but in a general manner.  
% \cite{acl/EbrahimiRLD18,cheng2018seq2sick,acl/SinghGR18} all worked on attacking  seq2seq (OpenNMT) \cite{acl/KleinKDSR17}
% \cite{conll/NiuB18} attacked DynoNet \cite{acl/HeBEL17}, which contains a Seq2seq-based utterance generator. 
% %
% \cite{corr/HeC2018} attacked different Seq2Seq models \cite{nips/SutskeverVL14,emnlp/ChoMGBBSB14}, while \cite{acl/WallaceB18} attacked  the Seq2Seq model proposed in \cite{emnlp/ChoMGBBSB14}. 

\subsubsection{Attention Models} 
Attention mechanism \cite{iclr/BahdanauCB15} is another breakthrough in deep leaning. 
It was initially developed to overcome the difficulty of encoding a long sequence required in Seq2Seq models \cite{springer/2018DengL}. 
Attention allows the decoder to look back on the hidden states of the source sequence. The hidden states then provide a weighted average as additional input to the decoder. This mechanism pays \textit{attention} on informative parts of the sequence. 
Rather than looking at the input sequence in vanilla attention models, self-attention  \cite{nips/VaswaniSPUJGKP17} in NLP is used to look at the surrounding words in a sequence  to obtain more contextually sensitive word representations \cite{cim/YoungHPC18}. 
BiDAF \cite{corr/SeoKFH16} is a bidirectional attention flow mechanism for machine comprehension and achieved outstanding performance when proposed. \cite{emnlp/JiaL17,acl/SinghGR18} evaluated the robustness of this model via adversarial examples and became the first few works using adversarial examples for attacking textual DNNs. % 
Other attention-based DNNs \cite{acl/Costa-JussaF16,emnlp/ParikhT0U16} also received adversarial attacks recently \cite{coling/EbrahimiLD18,conll/Minervini018}. 

%An attention-based Bi-RNN was proposed in \cite{acl/Costa-JussaF16} for neural machine translation. \cite{coling/EbrahimiLD18} perform adversarial attacks on this model. 
%
%\cite{emnlp/ParikhT0U16} proposed a  Decomposable Attention Model (DAM) for natural language inference. 
% \cite{conll/Minervini018} attacked  \cite{emnlp/ParikhT0U16}. 

% \cite{emnlp/JiaL17,acl/SinghGR18} attacked  BiDAF \cite{corr/SeoKFH16} which is a Bidirectional attention flow mechanism for machine comprehension. 
% %
%  \cite{coling/EbrahimiLD18} attacked attention-based Bi-RNN \cite{acl/Costa-JussaF16}. 
% %
% \cite{conll/Minervini018} attacked  Decomposable Attention Model (DAM) \cite{emnlp/ParikhT0U16}. 
%

\subsubsection{Reinforcement Learning Models} 
Reinforcement learning trains an agent by giving a reward after agents performing discrete actions. 
In NLP, reinforcement learning framework usually consist of an agent (a DNN), a policy (guiding action) and a reward. The agent picks an action (e.g., predicting next word in a sequence) based on a policy, then updates its internal state accordingly, until arriving the end of the sequence where a reward is calculated.
Reinforcement learning requires proper handling of the action and the states, which may limit the expressive power and learning capacity of the models \cite{cim/YoungHPC18}. But it gains much interests in task-oriented dialogue systems \cite{emnlp/LiMRJGG16} as they share the fundamental principle as decision making processes. Limited works so far can be found to attack the reinforcement learning model in NLP \cite{conll/NiuB18}.

\subsubsection{Deep Generative Models}
In recent years, two powerful deep generative models, Generative Adversarial Networks (GANs) \cite{nips/GoodfellowPMXWOCB14} and Variational Auto-Encoders (VAEs) \cite{iclr/KingmaW14} are proposed and gain much research attention. \cll{Generative models are able to generate realistic data that are very similar to ground truth data in a latent space.  In NLP field, they are used to generate textual data}. 
GANs \cite{nips/GoodfellowPMXWOCB14} consist of two adversarial networks: a generator and a discriminator. 
%\ah{(I prefer the definition of generator first and then Discriminator)} 
Discriminator is to discriminate the real and generated samples, while the generator is to generate realistic samples that aims to fool the discriminator. GAN uses a min-max loss function to train two neural networks simultaneously. 
VAEs consist of encoder and generator networks. Encoder  encodes an input into a latent space and the generator generates samples from the latent space. 
Deep generative models is not easy to train and evaluate. \cll{Hence, these deficiencies hinder their wide usage in many real-world applications}~\cite{cim/YoungHPC18}. Although they have been adopted in generating texts, so far no work examines their robustness using adversarial examples.  

\section{From  Image  to Text}
\label{sec:fromto}
Adversarial attacks are originated from computer vision community. In this section, we introduce representative works, discuss differences between attacking image data and textual data, and present preliminary knowledge when performing adversarial attacks on textual DNNs.

\subsection{Crafting Adversarial Examples: Inspiring Works in Computer Vision}
\label{sec:sub_cv}
Since adversarial examples are first proposed for attacking object recognition DNNs in computer vision community~ \cite{iclr14/Szegedy,iclr15/goodfellow,eurosp/PapernotMJFCS16, Dsp/Carlini017,cvpr/Moosavi-Dezfooli16,ccs/PapernotMGJCS17,zhao2017generating}, this research direction has been receiving sustained attentions. We briefly introduce some  works that inspired their followers in NLP community in this section, allowing the reader to better understand the adversarial attacks on textual DNNs. For comprehensive review of attack works in computer vision, please refer to \cite{access/AkhtarM18}.

% To solve Equation \ref{eq:adv}, the task is formulated as an optimization problem.  
% For an untargeted attack, the adversary is interested in any output that is different from the correct one. The problem is formulated as:

% \cll{
% \begin{eqnarray}
% \label{eq:untarget}
%  \max L(F(x'), y_x) \textrm{ s.t. } D(x,x')<\epsilon
% \end{eqnarray}
% }

% Maximizing the loss function would make the prediction goes to wrong direction. 
% %
% For targeted attack, the adversary has a targeted output $t$ and the optimization is:
% \cll{
% \begin{eqnarray}
% \label{eq:target}
%  \min L(F(x'), t) \textrm{ s.t. } D(x,x')<\epsilon
% \end{eqnarray}
% }
% %
% Minimizing the loss function given target output would enforce the model to give the target output. 
% %
%  Target attack is thus harder than untarget attack only if the problem is not a binary classification problem, for which target and untarget task are the same. 
 %
% Due to the non-convexity and non-linearity of NN, solving Eq. (\ref{eq:untarget}) and Eq. (\ref{eq:target}) are not always possible \cite{papernot2016crafting}. Thus current methods use approximation methods. We highlight some representative solutions that inspire many research efforts on generating textural adversarial examples. % 
 
%\att{from report: either change label or maximize loss} 
 
% 
\paragraph{L-BFGS}

Szegedy et al. invented the \textit{adversarial examples} notation \cite{iclr14/Szegedy}. They proposed a explicitly designed method to  cause the model to give wrong prediction of adversarial input ($\mathbf{x}+\mathbf{\eta}$) for image classification task. It came to solve the optimization problem:
\begin{eqnarray}
\label{eq:lbfgs}
\mathbf{\eta} = \arg \min_{\mathbf{\eta}} \lambda||\mathbf{\eta}||^2_2+J(\mathbf{x}+\mathbf{\eta},y')~~~ s.t.~~ (\mathbf{x}+\mathbf{\eta}) \in [0,1],
\end{eqnarray}
where $y'$ is the target output of ($\mathbf{x}'+\mathbf{\eta}$), but incorrect given an ideal classifier. $J$ denotes the cost function of the DNN and $\lambda$ is a hyperparameter to balance the two parts of the equation.  %\ah{(should refer to $\lambda$ what is it)} 
This minimization was initially performed with a box-constrained \textbf{L}imited memory \textbf{B}royden-\textbf{F}letcher-\textbf{G}oldfarb-\textbf{S}hanno (L-BFGS) algorithm and thus was named after it. The optimization was repeated multiple times until reaching a minimum $\lambda$ that satisfy Eq. (\ref{eq:lbfgs}). 

\paragraph{Fast Gradient Sign Method (FGSM) }
\label{sec:subsub_fgsm}
 L-BFGS is very effective, but highly expensive - this inspired Goodfellow et al. \cite{iclr15/goodfellow} to find a simplified solution. 
 Instead of fixing $y'$ and minimizing $\mathbf{\eta}$ in L-BFGS, FGSM fixed size of $\mathbf{\eta}$ and maximized the cost (Eq. (\ref{eq:FSGM})). Then they linearized the problem with a first-order Taylor series approximation (Eq. (\ref{eq:FSGM-li})), and got the closed-form solution of $\mathbf{\eta}$ (Eq. (\ref{eq:FSGM-cl})) \cite{warde201611}:
\begin{eqnarray}
\label{eq:FSGM}
&\mathbf{\eta} = \arg \max_{\mathbf{\eta}}J(\mathbf{x}+\mathbf{\eta},y)~~~ s.t.~~ ||\mathbf{\eta}||_\infty \leq \epsilon, \\
&\mathbf{\eta} = \arg \max_{\mathbf{\eta}}J(\mathbf{x},y)+\mathbf{\eta}^\mathbf{T}\nabla_{\mathbf{x}}J(\mathbf{x},y)~~~ s.t.~~ ||\mathbf{\eta}||_\infty \leq \epsilon, \label{eq:FSGM-li} \\
&\mathbf{\eta} = \epsilon\cdot \sign(\nabla_{\mathbf{x}}J(\mathbf{x},\mathbf{y})) \label{eq:FSGM-cl}
\end{eqnarray}
where  $\epsilon$ is a parameter set by attacker, controlling the perturbation's magnitude. \cll{sign(x) is the  sign function which returns 1 when $x>0$, and $-1$ when $x<0$, otherwise returns $0$}.  $\nabla_{\mathbf{x}}J(\mathbf{x},\mathbf{y})$ denotes the gradient of loss function respect to the input, and can be calculated via back-propagation.   %"Increasing the input variation parameter $\epsilon$ increases the likeliness of $\mathbf{x}'$ being misclassified but albeit simultaneously increases the perturbation’s magnitude and therefore its distinguishability."
FGSM attracts the most follow-up works in NLP. 
%If removing the sign operation, i.e., use the cost gradient to generate adversarial examples, the method comes to FGVM

\paragraph{Jacobian Saliency Map Adversary (JSMA)}
\label{sec:subsub_jsma}
Unlike FGSM using gradients to attack, Papernot et al. \cite{eurosp/PapernotMJFCS16} generated adversarial examples using forward derivatives (i.e., model Jacobian). This method evaluates the neural model's output sensitivity to each input component using its \textit{Jacobian Matrix} and gives greater control to adversaries given the perturbations.  Jacobian matrices form the adversarial saliency maps that rank each input component's contribution to the adversarial target. A perturbation is then selected from the maps.  Thus the method was named \textit{Jacobian-based Saliency Map Attack}.  
The Jacobian matrix of a given $\mathbf{x}$ is given by: 
\begin{eqnarray}
\label{eq:jsma}
Jacb_F[i,j] = \frac{\partial F_i}{\partial \mathbf{x}_j}
\end{eqnarray}
where $\mathbf{x}_i$ is the $i$-th component of the input and $F_j$ is the $j$-th component of the output. Here $F$ denotes the \textit{logits} (i.e., the inputs to the softmax  function) layer.  
$J_F[i,j]$ measures the sensitivity of $F_j$ with respect to  $\mathbf{x}_i$.

\paragraph{C\&W Attack}
\label{sec_subsub:cw}
Carlini and Wagner \cite{Dsp/Carlini017} aimed to evaluate the defensive distillation strategy \cite{corr/HintonVD15} for mitigating the adversarial attacks. They restricted the perturbations with $l_p$ norms where $p$ equals to $0, 2$ and $\infty$ and proposed seven versions of $J$ for the following optimization problem: 
\begin{eqnarray}
\label{eq:cw}
\mathbf{\eta} = \arg \min_{\mathbf{\eta}} ||\mathbf{\eta}||_p+ \lambda J(\mathbf{x}+\mathbf{\eta},y')~~~ s.t.~~ (\mathbf{x}+\mathbf{\eta}) \in [0,1],
\end{eqnarray}
and the formulation shares the same notation with aforementioned works. 
\paragraph{DeepFool}
DeepFool \cite{cvpr/Moosavi-Dezfooli16} is an iterative $L_2$-regularized algorithm. 
The authors first assumed the neural network is linear, thus they can separate the classes with a hyperplane. They simplified the problem and found optimal solution based on this assumption and constructed adversarial examples. To address the non-linearity fact of the neural network, they repeated the process until a true adversarial example is found.

\paragraph{Substitute Attack}
The above mentioned representative works are all white-box methods, which require the full knowledge of the neural model's parameters and structures. 
However, in practice, it is not always possible for attackers to craft adversaries in white-box manner due to the \cll{limited} access to the model. 
The limitation was addressed by Papernot et al. \cite{ccs/PapernotMGJCS17} and they introduced a black-box attack strategy: They trained a substitute model to approximate the decision boundaries of the target model with the labels obtained by querying the target model. Then they conducted white-box attack on this substitute and generate adversarial examples on the substitute. 
Specifically, they adopted FSGM and JSMA in generating adversarial examples for the substitute DNN.

\paragraph{GAN-like Attack}
There are another branch of black-box attack leverages the Generative Adversarial Neural (GAN) models.  Zhao et al. \cite{zhao2017generating} firstly trained a generative model, WGAN, on the training dataset $\mathbf{X}$. WGAN could generate data points that follows the same distribution with $\mathbf{X}$. Then they separately trained an inverter to map  data sample $\mathbf{x}$ to $\mathbf{z}$ in the latent dense space by minimizing the reconstruction error. 
Instead of perturbing $\mathbf{x}$, they searched for adversaries $z*$ in the neighbour of $\mathbf{z}$ in the latent space. Then they mapped $\mathbf{z*}$ back to $\mathbf{x*}$ and check if $\mathbf{x*}$ would change the prediction.  They introduced two search algorithms: iterative stochastic search and hybrid shrinking search. The former one used expanding strategy that gradually expand the search space, while the later one used shrinking strategy that \cll{starts} from a wide range and recursively \cll{tightens} the upper bound of the search range.

\subsection{Attacking Image DNNs vs Attacking Textual DNNs}
\label{sec:sub_diff}
To attack a textual DNN model, we cannot directly apply the approaches from the image DNN attackers as there are three main differences between them: 

\begin{itemize}
\item Discrete vs Continuous Inputs. 
Image inputs are continuous, typically the methods use $L_p$ norm measures the distance between clean data point with the perturbed data point. However, textual data is symbolic, thus discrete. It is hard to define the perturbations on texts. Carefully designed variants or distance measurements for textual perturbations are required. Another choice is to firstly map the textual data to continuous data, then adopt the attack method from computer vision. 
We will give further discussion in Section \ref{sec:sub_vectorizing}. 

\item Perceivable vs Unperceivable.
Small change of the image pixels usually can not be easily perceived by human beings, hence the adversarial examples will not change the human judgment, but only fool the DNN models. 
But small changes on texts, e.g., character or word change, will easily be perceived, rendering the possibility of attack failure. For example, the changes could be identified or corrected by spelling-check and grammar check before inputting into textual DNN models. 
Therefore, it is nontrivial to find  unperceivalble textual adversaries. 
\item Semantic vs Semantic-less. 
In the case of images, small changes usually do not change the semantics of the image as they are trivial and unperceivable. 
However, perturbation on texts would easily change the semantics of a word and a sentence, thus can be easily detected and heavily affect the model output. For example, deleting a negation word would change the sentiment of a sentence.  But this is not the case in computer vision where perturbing individual pixels does not turn the image from a cat to another animal.  
%
%However, the purpose of perturbation is to keep the correct prediction (usually by human) unchanged, but makes DNN model to be fooled to provide incorrect prediction. 
Changing semantics of the input is against the goal of adversarial attack that keep the correct prediction unchanged while fooling an victim DNN.  %\ah{(i didn't understand this phrases (fooling an vulnerable DNN))} .

\end{itemize}

Due to these differences, current \cll{state-of-the-art} textual DNN attackers either carefully adjust the methods from image DNN attackers by enforcing additional constraints, or propose novel methods using different techniques. 

\subsection{Vectorizing Textual Inputs and Perturbation Measurements}
\label{sec:sub_vectorizing}

\textbf{Vectorizing Textual Input.} DNN models \cll{require} vectors as input, for image tasks, the normal way is to use the pixel value to form the vectors/matrices as DNN input. But for \cll{textual} models, special operations are needed to transform the text into vectors. There are three main branches of methods: word-count based encoding, one-hot encoding and dense encoding (or feature embedding) and the later two are mostly used in DNN models of textual applications. % \att{Do I need to draw an image to illustrate the three encoding?}

\begin{itemize}
    \item \textit{Word-Count Based Encoding.}  Bag-of-words (BOW) method has the longest history in  vectorizing text. In BOW model, an zero-encoded vector with length of the vocabulary size is initialized. Then the dimension in vector is replaced by the count of corresponding word's appearance in the given sentence. Another word-count based encoding is to utilize the term frequency-inverse document frequency (TF-IDF) of a word (term), and the dimension in the vector is the TF-IDF value of the word.

    \item \textit{One-hot Encoding.} In one-hot encoding, a vector feature represents a token--a token could be a character (character-level model) or a word (word-level model). For character-level one-hot encoding, the representation can be formulated as \cite{acl/EbrahimiRLD18}:
    \begin{eqnarray}
    \mathbf{x}=[(x_{11},...x_{1n});...(x_{m1},...x_{mn})]\label{eq:onehotchar}
    \end{eqnarray}
    where $\mathbf{x}$ be a text of $L$ characters, $x_{ij} \in \{0,1\}^{|A|}$ and $|A|$ is the alphabet (in some works, $|A|$ also include symbols). 
    \cll{In Equation~\ref{eq:onehotchar}, $m$ is the number of words, $n$ is the maximum number of characters for a word in sequence $\textbf{x}$}. Thus each word has the same-fixed length of vector representation and the length is decided by the maximum number of characters of the words. 
    For word-level one-hot encoding, following the above notations, the text $x$ can be represented as:
    \begin{eqnarray}
    \mathbf{x}=[(x_1,...,x_m,x_{m+1}...x_k)]
    \end{eqnarray}
    where $x_{ij} \in \{0,1\}^{|V|}$ and $|V|$ is the vocabulary, which contains all words in a corpus. $k$ is the maximum number of words allowed for a text, so that $[(x_{m+1}...x_k)]$ is zero-paddings if $m+1<k$. 
    One-hot encoding produces vectors with only 0 and 1 values, where 1 indicates the corresponding character/word appears in the sentence/paragraph, while 0 indicate it does not appear.  Thus one-hot encoding usually generates sparse vectors/matrices.  
     DNNs have proven to be very successful in learning values from the sparse representations as they can learn more dense distributed representations from the one-hot vectors during the training procedure.
 
    \item \textit{Dense Encoding.} 
    Comparing to one-hot encoding, dense encoding generates low dimensional and distributed representations for textual data.  
    Word2Vec cite{nips/MikolovSCCD13} uses continuous bag-of-words (CBOW) and skip-gram models to generate dense representation for words, i.e., word embeddings. 
    It is based on the distributional assumption that words appearing within similar context possess similar meaning. % 
    Word embeddings, to some extend,  alleviates the discreteness and data-sparsity problems for vectorizing textual data \cite{synthesis/2017Goldberg}. 
    Extensions of word embeddings  such as doc2vec  and paragraph2vec \cite{icml/LeM14} encode sentences/paragraphs to dense vectors.  
\end{itemize}

\vspace{1mm} 
\textbf{Perturbation Measurement.} As described in Section \ref{sec:subsub_measure}, there needs a way to measure the size of the perturbation, so that it can be controlled to ensure the ability of fooling the victim DNN while remain unperceivable.  However, the measurement in textual perturbations is drastically different with the perturbations in image.  Usually, the size of the perturbation is measured by the distance between clean data $\mathbf{x}$ and its adversarial example $\mathbf{x'}$. But in texts, the distance measurement also need to consider the grammar correctness, syntax correctness and semantic-preservance. We here list the measurements used in the reviewed in this survey.

\begin{itemize}
\item \textit{Norm-based measurement.} Directly adopting norms such as $L_p, p\in {0,1,2,\infty} $ requires the input data are continuous. One solution is to use continuous and dense presentation (e.g., embedding) to represent the texts. But this usually results in invalid and incomprehensible texts, that need to involve other constrains.  

\item \textit{Grammar and syntax related measurement.} Ensuring the grammar or syntactic correctness makes the adversarial examples not easily perceived.  

\begin{itemize}
    \item \textit{Grammar and syntax checker} are used in some works to ensure the textual adversarial examples generated are valid. 
    \item \textit{Perplexity} is usually used to measure the quality of a language model. In one reviewed literature \cite{conll/Minervini018}, the authors used perplexity to ensure the generated adversarial examples (sentences) are valid. 
   \item \textit{Paraphrase} is controlled and can be regarded as a type of adversarial example (\ref{sec:subsub_paraphrase}). When perturbing, the validity of paraphrases is ensured in the generation process.   
\end{itemize}

\item \textit{Semantic-preserving measurement.} Measuring semantic similarity/distance is often performed on word vectors by adopting vectors' similarity/distance measurements. Given two $n$-dimensional word vectors 
$\mathbf{p}=(p_1, p_2,...,p_n)$ and $\mathbf{q}=(q_1, q_2,...,q_n)$: 

\begin{itemize}
\item \textit{Euclidean Distance} is a distance of two vectors  in the \textit{Euclidean} space:
\begin{eqnarray}
& d(\mathbf{p},\mathbf{q})=\sqrt{(p_1-q_1)^2+p_2-q_2)^2+..(p_n-q_n)^2}
\end{eqnarray}

\item \textit{Cosine Similarity} computes cosine value of the angle between the two vectors: 
\begin{eqnarray}
cos(\mathbf{p},\mathbf{q})=\frac{\sum_{i=1}^n p_i\times q_i}{\sqrt{\sum_{i=1}^n (p_i)^2}\times\sqrt{\sum_{i=1}^n (q_i)^2}}
\end{eqnarray}

\end{itemize}

\item  \textit{Edit-based measurement. }
Edit distance is a way of quantifying the minimum changes from one string to the other. Different definitions of edit distance use different sets of string operations \cite{ndss/LiJDLW19}. 

\begin{itemize}
    \item \textit{Levenshtein Distance} uses insertion, removal and substitution operations.
    \item  \textit{Word Mover's Distance (WMD)} \cite{icml/KusnerSKW15} is an edit distance operated on word embedding. It measures the minimum amount of distance that the embedded words of one document need to travel to reach the embedded words of the other document \cite{gong2018adversarial}. The minimization is formulated as:
    \begin{eqnarray}
   & \min \sum^n_{i,j=1}\textbf{T}_{ij}||\mathbf{e_i}-\mathbf{e_j}||_2 \\
  &  s.t., \sum^n_{j=1}\textbf{T}_{ij}=d_i, \forall i\in\{i,...,n\}, 
    \sum^n_{i=1}\textbf{T}_{ij}=d_i', \forall j\in\{i,...,n\} \nonumber
    \end{eqnarray}
    where $\mathbf{e_i}$ and $\mathbf{e_j}$ ared word embedding of word $i$ and word $j$ respectively. $n$ is the number of words. $\textbf{T}\in \mathcal{R}^{n\times n}$ be a flow matrix, where $\textbf{T}_{ij}\leq 0$ denotes how much of word $i$ in $\mathbf{d}$ travels to word $j$ in $\mathbf{d}'$. $\mathbf{d}$ and $\mathbf{d}'$ are normalized bag-of-words vectors of the source document and target document respectively. 

    \item   \textit{Number of changes} is a simple way to measure the edits and it is adopted in some reviewed literature. 
\end{itemize}

\item \textit{Jaccard similarity coefficient} is used for measuring similarity of finite sample sets utilising intersection and union of the sets.
\begin{eqnarray}
J(A,B) = \frac{|A\cap B|}{|A\cup B|}
\end{eqnarray}
In texts, $A$, $B$ are two documents (or sentences). $|A\cap B|$ denotes the number of words appear in both documents, $|A\cup B|$ refers to the number of unique words in total.

\end{itemize}

\section{Attacking Neural Models in NLP: State-of-The-Art}
\label{sec:attack}

In this section, we first introduce the categories of attack methods on textual  deep learning models
and then highlight the state-of-the-art research works, aiming to identify the most promising advances in recent years.

\subsection{Categories of Attack Methods on Textual Deep Learning Models}
\begin{figure}[!htb]
\centering
\includegraphics[width=0.9\textwidth]{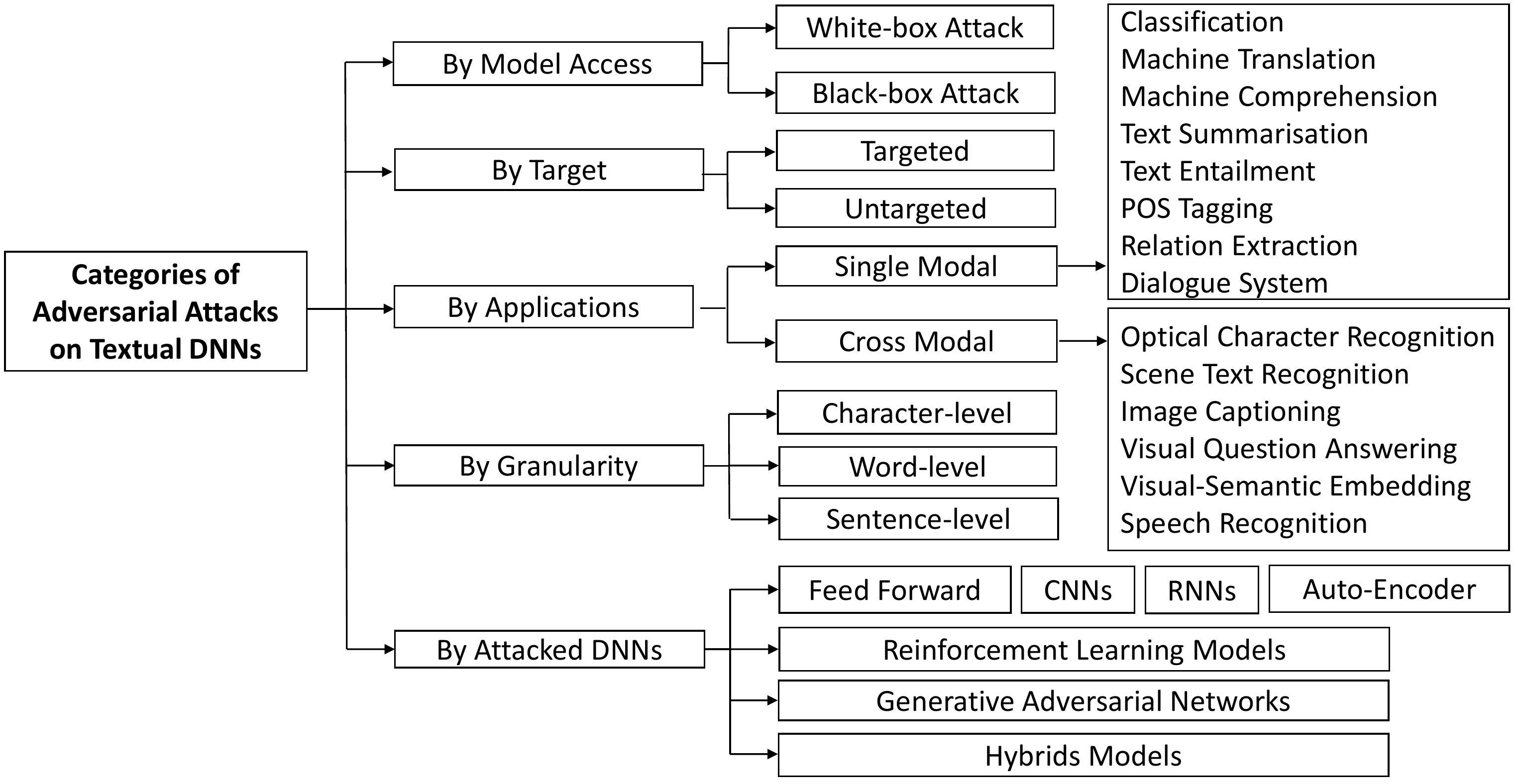}
\caption{Categories of Adversarial Attack Methods on Textual Deep Learning Models}
\label{fig:categories}
\end{figure}

We categorize existing adversarial attack methods based on different criteria. 
Figure \ref{fig:categories} generalizes the categories.   

In this article, five strategies are used to categorize the attack methods: i) \textit{By model access} group refers to the knowledge of attacked model when the attack is performed. In the following section, we focus on the discussion using this categorization strategy. ii)  \textit{By application} group refers the methods via different NLP applications. More detailed discussion will be provided in Section \ref{sec:data}.  iii)  \textit{By target} group  refers to the goal of the attack is enforcing incorrect prediction or targeting specific results. iv) \textit{By granularity} group considers on what granularity the model is attacked. v) We have discussed the \textit{attacked DNNs} in Section \ref{sec:sub_texNN}. In following sections, we will continuously provide information about different categories that the methods belong to.  

One important group of methods need to be noted is the cross-modal attacks, in which the attacked model consider the tasks dealing with multi-modal data, e.g., image and text data. They are not attacks for pure textual DNNs, hence we discuss this category of methods separately in Section \ref{sec:sub_multimodal} in addition to white-box attacks in Section \ref{sec:sub_white} and black-box attacks in Section \ref{sec:sub_black}.

\subsection{White-Box Attack}
\label{sec:sub_white}

In white-box attack, the attack requires the access to the model's full information, including architecture, parameters, loss functions, activation functions, input and output data. White-box attacks typically approximate the worst-case attack for a particular model and input, incorporating a set of perturbations.  This adversary strategy is often very effective. In this section, we group  white-box attacks on textual DNNs into seven categories.

\subsubsection{\cll{FGSM-based}}
\label{sec_subsub:fsgm_wb}
FGSM is one of the first attack methods on images (Section \ref{sec:subsub_fgsm}). It gains many follow-up works in attacking textual DNNs.   
TextFool \cite{liang2017deep} uses the concept of FGSM to approximate the contribution of text items that possess significant contribution to the text classification task. Instead of using sign of the cost gradient in FGSM, this work considers the magnitude. The authors proposed three  attacks: \textit{insertion}, \textit{modification} and \textit{removal}.  Specifically, they computed cost gradient $\Delta_xJ(f,x,c')$ of each training sample $x$, employing back propagation, where $f$ is the model function, $x$ is the original data sample, and $c'$ is the target text class. Then they identified the characters that contain the dimensions with the highest gradient magnitude and named them \textit{hot characters}.  Phrases that contain enough \textit{hot characters} and occur the most frequently are chosen as \textit{Hot Training Phrases} (HTPs). 
In the \textit{insertion} strategy, adversarial examples are crafted by inserting a few HTPs of the target class $c'$ nearby the phrases with significant contribution to the original class $c$. The authors further leveraged external sources like Wikipedia and \textit{forged fact} to select the valid and believable sentences. 
In the \textit{modification} Strategy, the authors identified \textit{Hot Sample Phrase (HSP)} to the current classification using similar way of identifying HTPs. Then they replaced the characters in HTPs by common misspellings or characters visually similar. 
In the \textit{removal} strategy, the inessential adjective or adverb in HSPs are removed.
The three strategies and their combinations are evaluated on a CNN text classifier  \cite{nips/ZhangZL15}. However, these methods are performed manually, as mentioned by the authors.  

%
%The works \cite{samanta2017towards} and \cite{ecir/SamantaAdversarial} 
The work in \cite{ecir/SamantaAdversarial} adopted the same idea as TextFool, but it provides a \textit{removal}-\textit{addition}-\textit{replacement} strategy that firstly tries to \textit{remove} the adverb ($w_i$) which contributed the most to the text classification task (measured using loss gradient). If the output sentences in this step have incorrect grammar, the method will \textit{insert} a word $p_j$ before $w_i$. $p_j$ is selected from a candidate pool,  in which the synonyms and typos and genre specific keywords (identified via term frequency) are candidate words. 
If the output cannot satisfy the highest cost gradient for all the $p_j$, then the method \textit{replaces} $w_i$ with $p_j$. The authors showed that their method is more effective than TextFool.     
As the method ordered the words with their contribution ranking and crafted adversarial samples according to the order, it is a greedy method that always get the minimum manipulation until the output changes. To avoid being detected by the human eyes, the authors constrained the replaced/added words to not affect the grammar and POS of the original \cll{words}.

In malware detection, an portable executable (PE) is represented by binary vector $\{x_1,...,x_m\}$, $x_i\in\{0,1\}$ that using 1 and 0 to indicate the PE is present or not where $m$ is the number of PEs. Using  PEs' vectors as features, malware detection DNNs can identify the malicious software. 
It is not a typical textual application, but also targets discrete data, which share similar methods with textual applications. 
The authors of  \cite{huang2018adversarial}  investigated the methods to generate binary-encoded adversarial examples. 
To preserve the functionality of the adversarial examples, they incorporated four bounding methods to craft perturbations. 
The first two methods adopt FSGM$^k$ \cite{iclr/KurakinGB17}, the multi-step variant of FGSM, restricting the perturbations in a binary domain by introducing deterministic rounding (dFGSM$^k$) and randomized rounding (rFGSM$^k$). These two bounding methods are similar to $L_\infty$-ball constraints on images \cite{iclr15/goodfellow}.  
The third method multi-step Bit Gradient Ascent (BGA$^k$) sets the bit of the $j$-th feature if the corresponding partial derivative of the loss is greater than or equal to the loss gradient's $L_2$-norm divided by $\sqrt{m}$. 
The fourth method multi-step Bit Coordinate Ascent (BCA$^k$) updates one bit in each step by considering the feature with the maximum corresponding partial derivative of the loss. 
These two last methods actually visit multiple feasible vertices. 
The work also proposed a adversarial learning framework aims to robustify the malware detection model. 
% 
%"Equipping projected gradient descent with randomness in rounding helped uncover roughly 4 times as many malicious samples in the binary feature space as those uncovered in natural training"

% \cite{eusipco/KolosnjajiDBMGE18} also attacks malware detection    use  gradient

\cite{rosenberg2017generic} also attacks malware detection DNNs. The authors made perturbations on the embedding presentation of the binary sequences and reconstructed the perturbed examples to its binary representation. 
%In order to preserve the malicious functionality of the malware, that is, the adversarial examples is incorrectly labelled as benigh by the victim DNN detector, the authors injected small chunk of bytes (payload) to the original inputs. 
Particularly, they appended a uniformly random sequence of bytes (payload) to the original binary sequence. Then they embed the new binary to its embedding and performed FGSM only on the embedding of the payload. The perturbation is performed iteratively until the detector output incorrect prediction. Since the perturbation is only performed on payload, instead of the input, this method will preserve the functionality of the malware. 
Finally, they reconstructed adverse embedding to valid binary file by mapping the adversary embedding to its closest neighbour in the valid embedding space.

% adv training use FGSM
%
Many works directly adopt FGSM for adversarial training, i.e., put it as regularizer when training the model.  We will discuss some representatives in Section \ref{sec:defense}.

\subsubsection{JSMA-based}
JSMA is another pioneer work on attacking neural models for image applications (refers to Section \ref{sec:subsub_jsma}). 
The work \cite{papernot2016crafting} used forward derivative as JSMA to find the most contributable sequence towards the adversary direction.  The network's Jacobian had been calculated by leveraging computational graph unfolding \cite{mozer1995focused}. 
They crafted adversarial sequences for two types of RNN models whose output is categorical and sequential data respectively. 
For categorical RNN, the adversarial examples are generated by considering the Jacobian $Jacb_F[:,j]$ column corresponding to one of the output components $j$. Specifically, for each word $i$, they identified the direction of perturbation by: 
\begin{eqnarray}
\label{eq:jasm1}
& \sign(Jacb_F(x')[i,g(x')]) \\
& g(x') = \arg \max_{0,1}(p_j)
\end{eqnarray}
where $p_j$ is the output probability of the target class. As in JSMA, they instead to choose logit to replace probability in this equation. They further  projected the perturbed examples onto the closest vector in the embedding space to get valid embedding. 
For sequential RNN, after computing the Jacobian matrix, they altered the subset of input setps $\{i\}$ with high Jacobian values $Jacb_F[i,j]$ and low Jacobian values $Jacb_F[i,k]$ for $k  \neq j$ to achieve modification on a subset of output steps $\{j\}$.

\cite{grosse2016adversarial} (and \cite{ESORICS17/GrosseAdvMalware}) is the first work to attack neural malware detector.  They firstly performed feature engineering and obtained more than 545K static features for software applications. They used binary indicator feature vector to represent an application.  
Then they crafted adversarial examples on the input feature vectors by adopting JSMA:  they computed gradient of model Jacobian to estimate the perturbation direction. 
Later, the method chooses a perturbation $\eta$ given input sample that with maximal positive gradient into the target class. In particular, the perturbations are chosen via index $i$, satisfying:
\begin{eqnarray}
\label{eq:jasm2}
& i = \arg \max_{j\in[1,m],\mathbf{X}_j=y'} f_y'(\mathbf{X}_j)
\end{eqnarray}
where $y'$ is the target class, $m$ is the number of features. On the binary feature vectors, the perturbations are ($0\rightarrow 1$) or ($1\rightarrow 0$). This method preserves the functionality of the applications.   In order to ensure that modifications caused by the perturbations do not change the application much, which will keep the malware application's functionality complete, the authors used the $L_1$ norm to bound the overall number of features modified, and further bound the number of features to 20. 
In addition, the authors  provided three methods to \cll{defense} against the attacks, namely feature reduction, distillation and adversarial training. They  found adversarial training is the most effective defense method.

\subsubsection{C\&W-based}
% %
% C\&W method is an optimization-based method that turns the adversarial example generation into an optimization problem. 
% Generally, the problem can be formulated as: 
% \begin{eqnarray}
% \label{eq:opts}
% \min_\delta L(X+\delta) + \lambda R(\delta)
% \end{eqnarray}
% where $L(\cdot)$ is the loss function to penalize the unsuccessful attack,  $R(\cdot)$ indicates the regularization function to measure the magnitude of distortions. $\lambda $ is the regularization parameter that control the trade-off between success attack and distortion. 
% %

The work in \cite{kdd/SunTYWZ18} adopted C\&W method (refer\ah{s} to Section \ref{sec:sub_cv}) for attacking predictive models of medical records. The aim is to detect susceptible events and measurements in each patient's medical records, which provide guidance for the clinical usage. The authors used standard LSTM as predictive model. Given the patient EHR data \cll{being} presented by a matrix $X^i\in\mathbf{R}^{d\times t_i}$ ($d$ is the number of medical features and $t_i$ is the time index of medical check), the generation of the adversarial example is formulated as:
\begin{eqnarray}
\label{eq:cw1}
\min_{\hat{X}}\max\{-\epsilon, [logit(\mathbf{x'})]_{y}-[logit(\mathbf{x})]_{y'} \} + \lambda||\mathbf{x'}-\mathbf{x}||_1
\end{eqnarray}
where $logit(\cdot)$ denotes the logit layer output, $\lambda$ is the regularization parameter which controls the $L_1$ norm regularization, $y'$ is the targeted label while $y$ is the original label. 
%$\hat{X}$ is the adversarial example of $X$. 
After generating adversarial examples, the authors picked the optimal example according to their proposed evaluation scheme that considers both the perturbation magnitude and the structure of the attacks.  
Finally they  used the adversarial example to compute the susceptibility
score for the EHR as well as the cumulative susceptibility score for
different measurements.

Seq2Sick \cite{cheng2018seq2sick} attacked the seq2seq models using two targeted attacks: non-overlapping attack and keywords attack. For non-overlapping attack, the authors aimed to generate adversarial sequences that are entirely different from the original outputs. They proposed a hinge-like loss function that optimizes on the logit layer of the neural network: 
\begin{eqnarray}
\label{eq:hinge}
\sum^{|K|}_{i=1} \min_{t\in[M]}\{m_t(\max\{-\epsilon,\max_{y\neq k_i}\{z_t^{(y)}\}-z_t^{(k_i)} \})  \}
\end{eqnarray}
where $\{s_t\}$ are the original output sequence, $\{z_t\}$ indicates the logit layer outputs of the adversarial example. 
For the keyword attack, targeted keywords are expected to appear in the output sequence. The authors also put the optimization on the logit layer and tried to ensure that the targeted keyword's logit be the largest among all words. Further more,  they defined mask function $m$ to solve the keyword collision problem. The loss function then becomes:
\begin{eqnarray}
\label{eq:hinge}
L_{keywords} = \sum^{|K|}_{i=1} \min_{t\in[M]}\{m_t(\max\{-\epsilon,\max_{y\neq k_i}\{z_t^{(y)}\}-z_t^{(k_i)} \})  \}
\end{eqnarray}
where $k_i$ denotes the $i$-th word in output vocabulary. 
To \cll{ensure} the generated word embedding is valid, this work also considers two regularization methods: group lasso regularization to enforce the group sparsity, and group gradient regularization to make adversaries are in the permissible region of the embedding space.

\subsubsection{Direction-based}
HotFlip \cite{acl/EbrahimiRLD18} performs atomic flip operations to generate adversarial examples. Instead of leveraging gradient of loss, HotFlip use the directional derivatives. Specifically, HotFlip represents character-level operations, i.e., swap, insert and delete, as vectors in the input space and estimated the change in loss by directional derivatives with respect to these vectors. 
Specifically, given one-hot representation of inputs, a character flip in the $j$-th character of the $i$-th word (a$\rightarrow$b) can be represented by the  vector:
\begin{eqnarray}
\label{eq:hfv}
& \overrightarrow{v}_{ijb}= (\mathbf{0} ,..;(\mathbf{0} ,..(0,..-1,0,..,1,0)_j ,..\mathbf{0} )_i; \mathbf{0} ,..)
\end{eqnarray}
where -1 and 1 are in the corresponding positions for the a-th and b-th characters of the alphabet, respectively. 
Then the best character swap can be found by maximizing a first-order approximation of loss change  via directional derivative along the operation vector: 
\begin{eqnarray}
\label{eq:hotflip}
\max \nabla_x J(x,y)^T \cdot \overrightarrow{v}_{ijb}=\max_{ijv}\frac{\partial J^{(b)}}{\partial x_{ij}}-\frac{\partial J^{(a)}}{\partial x_{ij}}
\end{eqnarray}
where $J(x,y)$ is the model's loss function with input $x$ and true output $y$.
Similarly, insertion at the $j$-th position of the $i$-th word can also be treated as a character flip, followed by more flips as characters are shifted to
the right until the end of the word. The character deletion is a 
number of character flips as characters are shifted to the left. 
 Using the beam search, HotFlip efficiently finds the best directions for multiple flips.  

The work \cite{coling/EbrahimiLD18} extended HotFlip by adding targeted attacks. Besides the swap, insertion and deletion as provided in HotFlip, the authors proposed a controlled attack, which is to remove a specific word from the output, and a targeted attack, which is to replace a specific word by a chosen one. 
To achieve these attacks, they maximized the loss function $J(x,y_t)$ and minimize   $J(x,y_t')$, where $t$ is the target word for the controlled attack, and $t'$ is the word to replace $t$. 
Further, they proposed three types of attacks that provide multiple modifications. In \textit{one-hot} attack, they manipulated all the words in the text with the best operation. In \textit{Greedy} attack, they make another forward and backward pass, in addition to picking the best operation from the whole text. 
In \textit{Beam search} attack, they replaced the search method in greedy with the beam search. 
In all the attacks proposed in this work, the authors set threshold for the maximum number of changes, e.g., 20\% of characters are allowed to be changed.

\begin{table}[t]\footnotesize
\begin{tabular}{|l|l|p{20mm}|c|p{22mm}|p{28mm}|p{8mm}|}
\hline
\textbf{Strategy}  & \textbf{Work}  & \textbf{Granularity} & \textbf{Target}  & \textbf{Attacked Models}  &  \textbf{Perturb Ctrl.}   &  \textbf{App.}    \\ \hline \hline
\multirow{7}{*}{FSGM-based} & \cite{liang2017deep}  & character,word  & Y   & CNN  \cite{nips/ZhangZL15} & $L_\infty$&  TC \\ 
%
%\cline{2-7} & \cite{samanta2017towards}         & word            & N    &  &  &   \\ 

\cline{2-7} & \cite{ecir/SamantaAdversarial}  & word  & N  & CNN  \cite{nips/ZhangZL15} & $L_\infty$, Grammar and POS correctness &  TC \\  
%\cline{2-7}  & \cite{wu2017adversarial}          & sentence        & N    &  & &     \\ 
\cline{2-7}  & \cite{rosenberg2017generic}       & PE     & binary      & CNN in \cite{icassp/DahlSDY13}  & Boundaries employ $L_\infty$ and $L_2$  & MAD   \\ 
\cline{2-7}  & \cite{huang2018adversarial}    & PE embedding   & binary & MalConv  \cite{aaai/RaffBSBCN18} & $L_\infty$  &  MAD  \\
\hline
 \multirow{2}{*}{JSMA-based}  & \cite{papernot2016crafting}       & word embedding    & binary  &  LSTM & -- & TC \\ 
\cline{2-7} & \cite{grosse2016adversarial,ESORICS17/GrosseAdvMalware}      & application features     & binary   & Feed forward &  $L_1$ & MAD\\ 
\hline
\multirow{2}{*}{C\&W-based }   & \cite{kdd/SunTYWZ18}      & medical features & Y    & LSTM  & $L_1$  &  MSP   \\ 
\cline{2-7} & \cite{cheng2018seq2sick}          & word embedding       & Y    &  OpenNMT-py \cite{acl/KleinKDSR17} & $L_2$+gradient regularization &  TS, MT   \\ \hline
\multirow{2}{*}{Direction-based}     & \cite{acl/EbrahimiRLD18}        & character       & N    & CharCNN-LSTM \cite{aaai/KimJSR16} & -- & TC \\ 
\cline{2-7} & \cite{coling/EbrahimiLD18}        & character       & Y & CharCNN-LSTM \cite{acl/Costa-JussaF16}  & Number of changes & MT \\ \hline
Attention-based   & \cite{conll/BlohmJSYV18}          & word, sentence  & N  & \cite{wang2016compare,corr/abs-1709-05036,dzendzik2017framed}, CNN, LSTM and  ensembles  &  Number of changes &  MRC, QA \\  \hline
Reprogramming   & \cite{corr/NeekharaHDK2018}          &  word &  N & CNN, LSTM, Bi-LSTM  &  -- &  TC \\  \hline
Hybrid   & \cite{gong2018adversarial}        & word embedding       & N    &  CNN &  WMD & TC, SA  \\ \hline
\end{tabular}
\caption{Summary of reviewed white-box attack methods. PE: portable executable; TC: text classification; SA: sentiment analysis; TS: text summarisation; MT: machine translation MAD: malware detection; MSP: Medical Status Prediction; MRC: machine reading comprehension; QA: question answering; WMD: Word Mover's Distance; --: not available.}
\label{tbl:white}
\end{table}

\subsubsection{Attention-based}
 \cite{conll/BlohmJSYV18} proposed two white-box attacks for the purpose of comparing the robustness of CNN verses RNN. 
 They leveraged the model's internal attention distribution to find the \cll{pivotal sentence which is assigned a larger weight by  the model to derive the correct answer}. Then they exchanged the words which received the most attention with the randomly chosen words in a known vocabulary. They also performed another white-box attack by removing the whole  sentence that gets the highest attention. Although they focused on attention-based models, their attacks do not examine the attention mechanism itself, but solely leverages the outputs of the attention component (i.e., attention score). 

\subsubsection{Reprogramming} 
\cite{corr/NeekharaHDK2018} adopts adversarial reprogramming (AP) to attack sequence neural classifiers. 
AP \cite{corr/abs-1806-11146} is a recently proposed adversarial attack where a adversarial reprogramming function $g_\theta$ is trained to  re-purpose the attacked DNN to perform a alternate task (e.g., question classification to name classification) without modifying the DNN's parameters. AP adopts idea from transfer learning, but keeps the parameters unchanged. 
The authors in \cite{corr/NeekharaHDK2018} proposed both white-box and black-box attacks. In white-box, Gumbel-Softmax is applied to train $g_\theta$ who can work on discrete data. We discuss the black-box method later. They evaluated their methods on various text classification tasks and confirmed the effectiveness of their methods.

\subsubsection{Hybrid}
Authors of the work \cite{gong2018adversarial} perturbed the input text on word embedding against the CNN model. This is a general method that is applicable to most of the attack methods developed for computer vision DNNs. The authors specifically applied FGSM and DeepFool. %We consider this method as \textit{hybrid of methods}. %
Directly applying methods from computer vision would generate meaningless adversarial examples. To address this issue, the authors rounded the  adversarial examples to the nearest meaningful word vectors by using Word Mover's Distance (WMD) as the distance measurements. The evaluations on sentiment analysis and text classification datasets show that WMD is a qualified metric for controlling the perturbations.

\vspace{2mm}
\noindent \textbf{Summary of White-box Attack.}
We summarize the reviewed white-box attack works in Table \ref{tbl:white}. We highlight four aspects include granularity-on which level the attack is performed; target-whether the method is  target or un-target; the attacked model, perturbation control-methods to control the size of the perturbation, and applications. It is worth noting that in binary classifications, target and untarget methods show same effect, so we point out their \textit{target} as ''binary" in the table.

\subsection{Black-box Attack}
\label{sec:sub_black}
Black-box attack does not require the details of the neural networks, but can access the input and output. This type of  attacks often rely on heuristics to generate adversarial examples, and it is  more practical as in many \cll{real-world} applications the details of the DNN is a black box to the attacker. In this article, we group black-box attacks on textual DNNs into five categories.

\subsubsection{Concatenation Adversaries}
\label{sec:_subsub_concat}
\begin{figure}[b]
\centering
\includegraphics[width=0.5\textwidth]{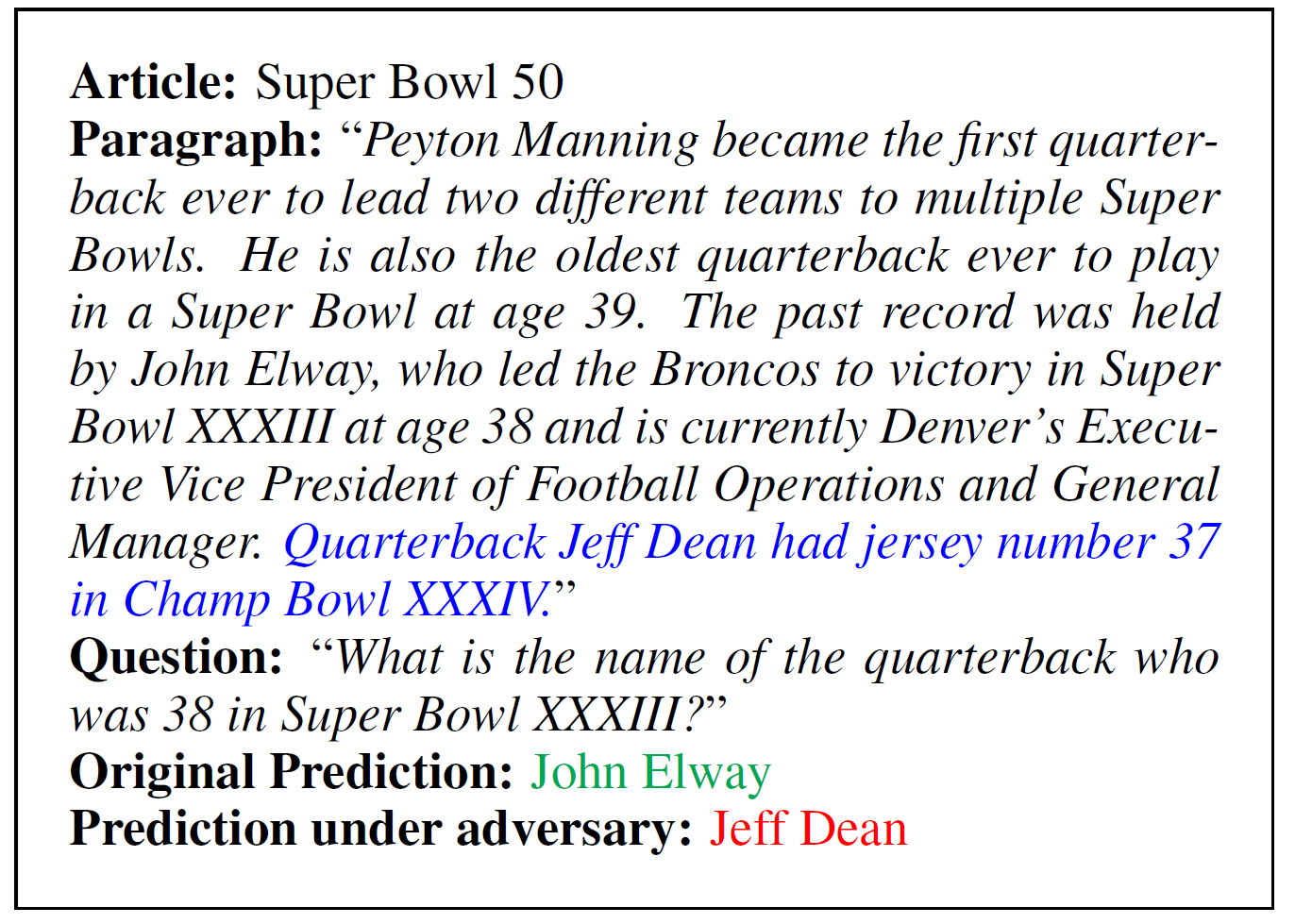}
\caption{Concatenation adversarial attack on reading comprehension DNN. After adding distracting sentences (in blue) the answer changes from correct one (green) to incorrect one (red) \cite{emnlp/JiaL17}.}
\label{fig:bl-can}
\end{figure}
\begin{figure}[b]
\centering
\includegraphics[width=0.6\textwidth]{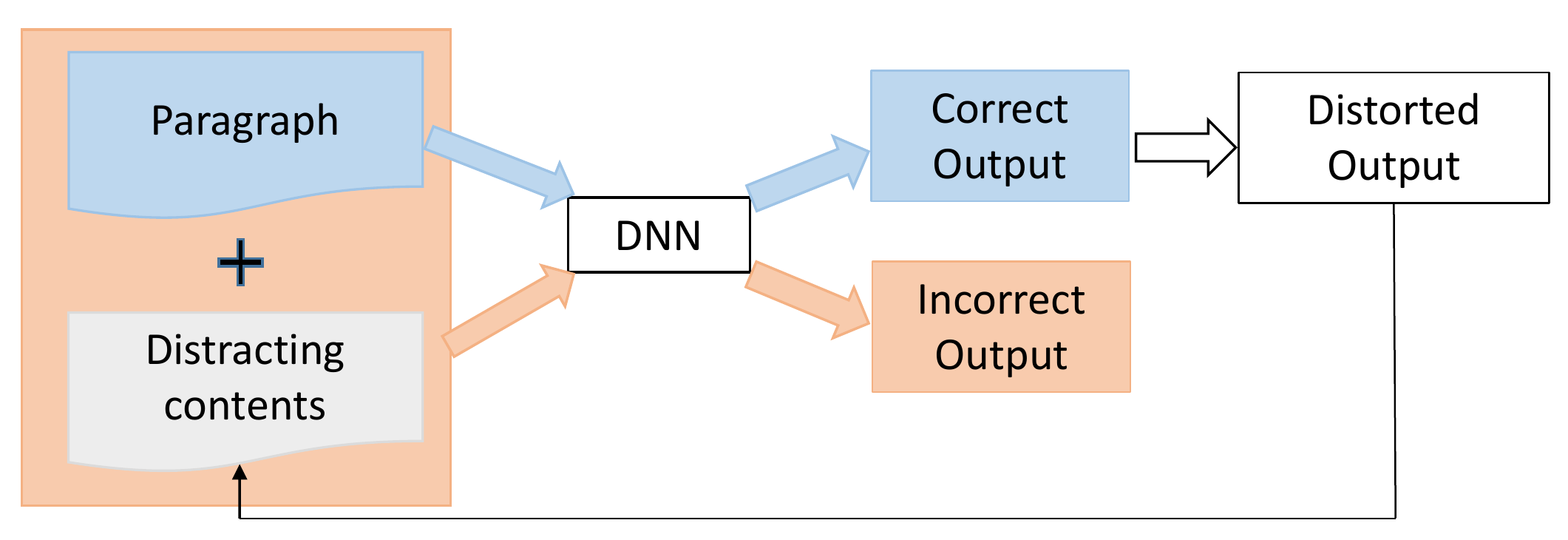}
\caption{General principle of concatenation adversaries. Correct output are often utilized to generate distorted output, which later will be used to build distracting contents.  Appending distracting contents to the original paragraph as adversarial input to the attacked DNN and cause the attacked DNN produce incorrect output. }
\label{fig:bl-can-wf}
\end{figure}

 \cite{emnlp/JiaL17} is the first work to attack reading comprehension systems. The authors proposed \textit{concatenation adversaries}, \cll{which} is to append distracting but meaningless sentences at the end of the paragraph. \cll{These distracting sentences} do not change the semantics of the paragraph and the question answers, but will fool the neural model. The distracting sentences are either carefully-generated \cll{informative} sentences or arbitrary sequence of words using a pool of 20 random common words.  
 Both perturbations were obtained by iteratively querying the neural network until the output changes.  Figure \ref{fig:bl-can} illustrates an example from \cite{emnlp/JiaL17} that after adding distracting sentences (in blue) the answer changes from correct one (green) to incorrect one (red). 
The authors of \cite{naacl/WangM_naacl18}  improved the work by varying the locations where the distracting sentences are placed and expanding the set of fake answers for generating the distracting sentences, rendering new adversarial examples that can help training more robust neural models.
 \cll{Also, the work \cite{conll/BlohmJSYV18} utilized the distracting sentences to evaluate the robustness of their reading comprehension model. Specifically, they use a pool of ten random common words in conjunction with all question words and the words from all incorrect answer candidates to generate the distracting sentences}. In this work, a simple word-level black-box attack is also performed by replacing the most frequent words via their synonyms. As aforementioned, the authors also provided two white-box strategies. 
 Figure \ref{fig:bl-can-wf} illustrates the general workflow for concatenation attack. Correct output (i.e., answer in MRC tasks) are often leveraged to generate distorted output, which later will be used to build distracting contents.  Appending distracting contents to the original paragraph as adversarial input to the attacked DNN. The distracting contents will not \textit{distract} human being and ideal DNNs, but can make vulunerable DNNs to produce incorrect output.

% \begin{figure}[]
%   \centering
%   \subfigure[Example in \cite{emnlp/JiaL17}]{
%     \label{fig:bl-pliang} 
% 	\includegraphics[width=0.4\linewidth]{figs/black-can.png}} 
%   % \hspace{-3mm}
%   \subfigure[The General Workflow of Concatenate Strategy]{
%      \label{fig:bl-cam} 
% 	\includegraphics[width=0.5\linewidth]{figs/black-can-crop}} 
%   \caption{Concatenate Adversarial Attacks. }
%   \label{fig:synnoises} 
% \end{figure}

\subsubsection{Edit Adversaries}
\label{sssec:editAdv} 

\begin{figure}[t]
\centering
\includegraphics[width=0.9\textwidth]{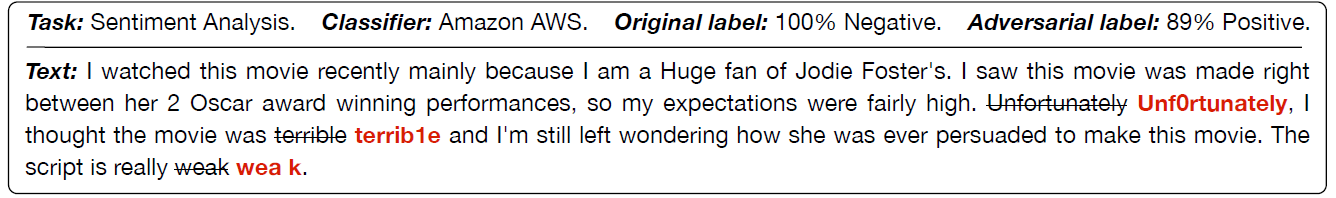}
\caption{Edit adversarial attack on sentiment analysis DNN. After editing words (red), the prediction changes from 100\% of Negative to 89\% of Positive \cite{ndss/LiJDLW19}.}
\label{fig:bl-edit}
\end{figure}
\begin{figure}[b]
\centering
\includegraphics[width=0.5\textwidth]{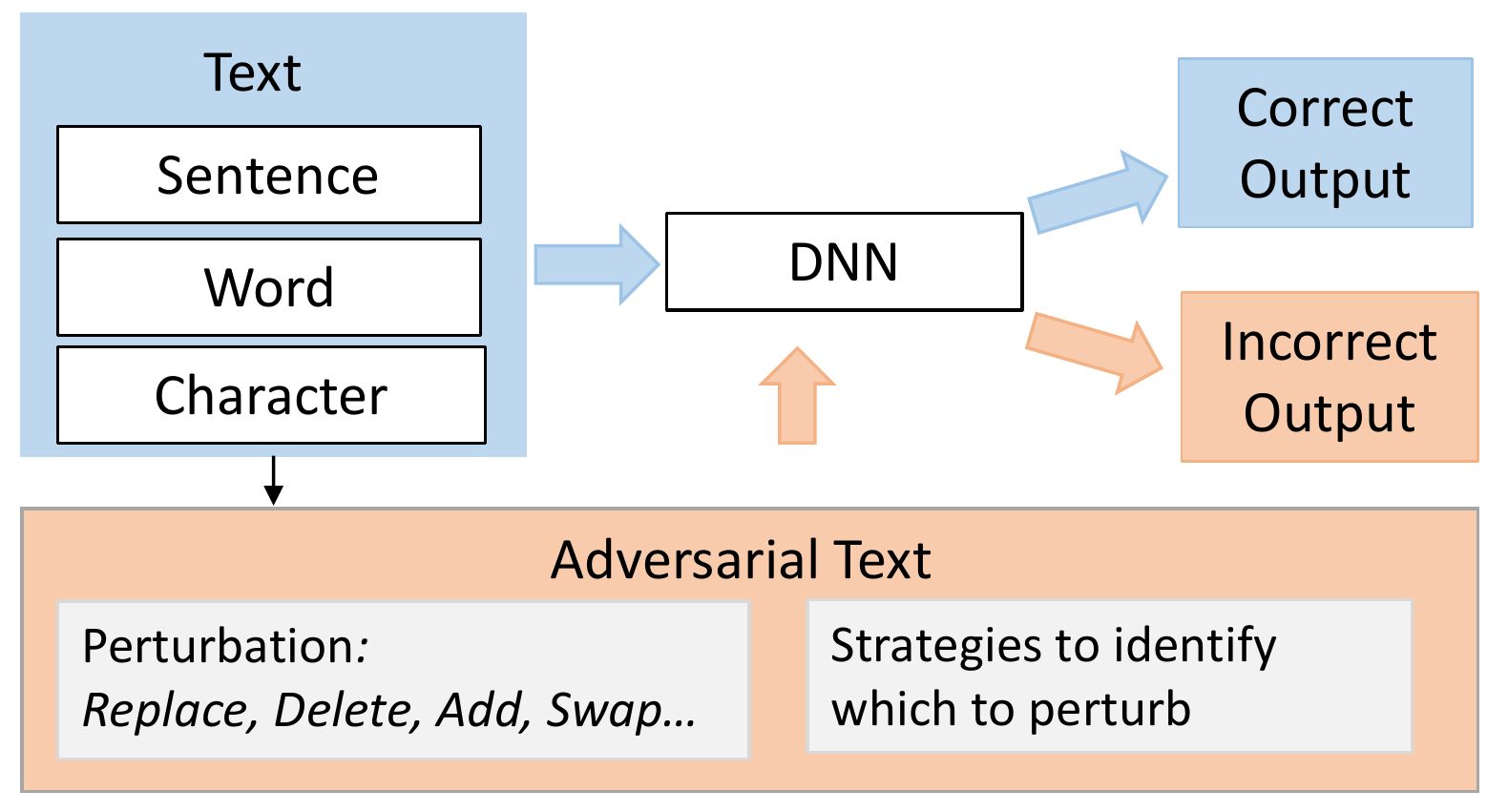}
\caption{General principle of edit adversaries. Perturbations are performed on sentences, words or characters by edit strategies such as replace, delete, add and swap. }
\label{fig:bl-edit-wf}
\end{figure}
The work in \cite{belinkov2017synthetic}  perturbed the input data of neural machine translation applications in two ways: 
\textit{Synthetic},  which performed the character order changes, such as swap, middle random (i.e., randomly change orders of characters except the first and the last), fully random (i.e., randomly change orders of all characters) and keyboard type. They also collected typos and misspellings as adversaries. \textit{natural}, leveraged the typos from the datasets. 
Furthermore, \cite{conll/NiuB18} attacked the neural models for dialogue generation. They applied various perturbations in  dialogue context, namely Random Swap (randomly transposing neighboring tokens) and Stopword Dropout (randomly removing stopwords), Paraphrasing (replacing words with their paraphrases),  Grammar Errors (e.g., changing a verb to the wrong tense) for the \textit{Should-Not-Change} attacks, and the Add Negation
strategy (negates the root verb of the source input) and  Antonym strategy (changes verbs, adjectives, or adverbs to their antonyms) for \textit{Should-Change} attacks.  
DeepWordBug \cite{gao2018black} is a simple method that uses character transformations to generate adversarial examples. The authors first identified the important `tokens', i.e., words or characters that affect the model prediction the most by scoring functions developed by measuring the DNN classifier's output. Then they modified the identified tokens using four strategies: replace, delete, add and swap. The authors evaluated their method on a variety of NLP tasks, e.g., text classification, sentiment analysis and spam detection. 
\cite{ndss/LiJDLW19} followed \cite{gao2018black}, refining the scoring function. Also this work provided white-box attack adopting JSMA. One contribution of this work lies on the perturbations are restricted using four textual similarity measurement: edit distance of text; Jaccard similarity coefficient; Euclidean distance on word vector; and cosine similarity on word embedding. Their method had been evaluated only on sentiment analysis task.

The authors in \cite{conll/Minervini018} proposed a method for  automatically generating adversarial examples that violate a set of given First-Order Logic constraints in natural language inference (NLI). They proposed \cll{an} inconsistency loss to measure the  degree to which a set of sentences causes a model to violate a rule. The adversarial example generation is the process for finding the mapping between variables in rules to sentences that maximize the inconsistency loss and  are composed by sentences with a low perplexity (defined by a language model). To generate  low-perplexity adversarial sentence examples, they used three edit perturbations: i) change one word in one of the input sentences; i) remove one parse subtree from one of the input sentences; iii) insert one parse sub-tree from one sentence in the corpus in
the parse tree of \cll{the another sentence}.

The work in \cite{emnlp/AlzantotSEHSC18} uses genetic algorithm (GA) for minimising the number of word replacement from the original text, but at the same time can change the result of the attacked model. They adopted \textit{crossover} and \textit{mutation} operations in GA to generate perturbations. The authors measured the effectiveness of the word replacement accoding to the impact on attacked DNNs. Their attack focused on sentiment analysis and textual entailment DNNs. 

In \cite{corr/ChanMXXLO2018}, the authors proposed a framerwork for adversarial attack on Differentiable Neural Computer (DNC).
DNC is a computing machine with DNN as its central controller
operating on an external memory module for data processing. 
Their method uses two new automated and scalable strategies to generate grammatically corret adversairal attacks in question answering domian, utilising metamorphic transformation. The first strategy, \textit{Pick-n-Plug}, consists of   a pick operator pick to draw adversarial sentences from a particular task (source task) and plug operator plug to inject these sentences into a story from another task (target task), without changing its correct answers.
Another strategy, \textit{Pick-Permute-Plug}, extends the adversarial capability of P\textit{Pick-n-Plug} by an additional permute operator after picking sentences (gpick) from a source task. Words in a particular adversarial sentence can be permuted with its synonyms to generate a wider range of possible attacks.

\begin{figure}[t]
\centering
\includegraphics[width=0.5\textwidth]{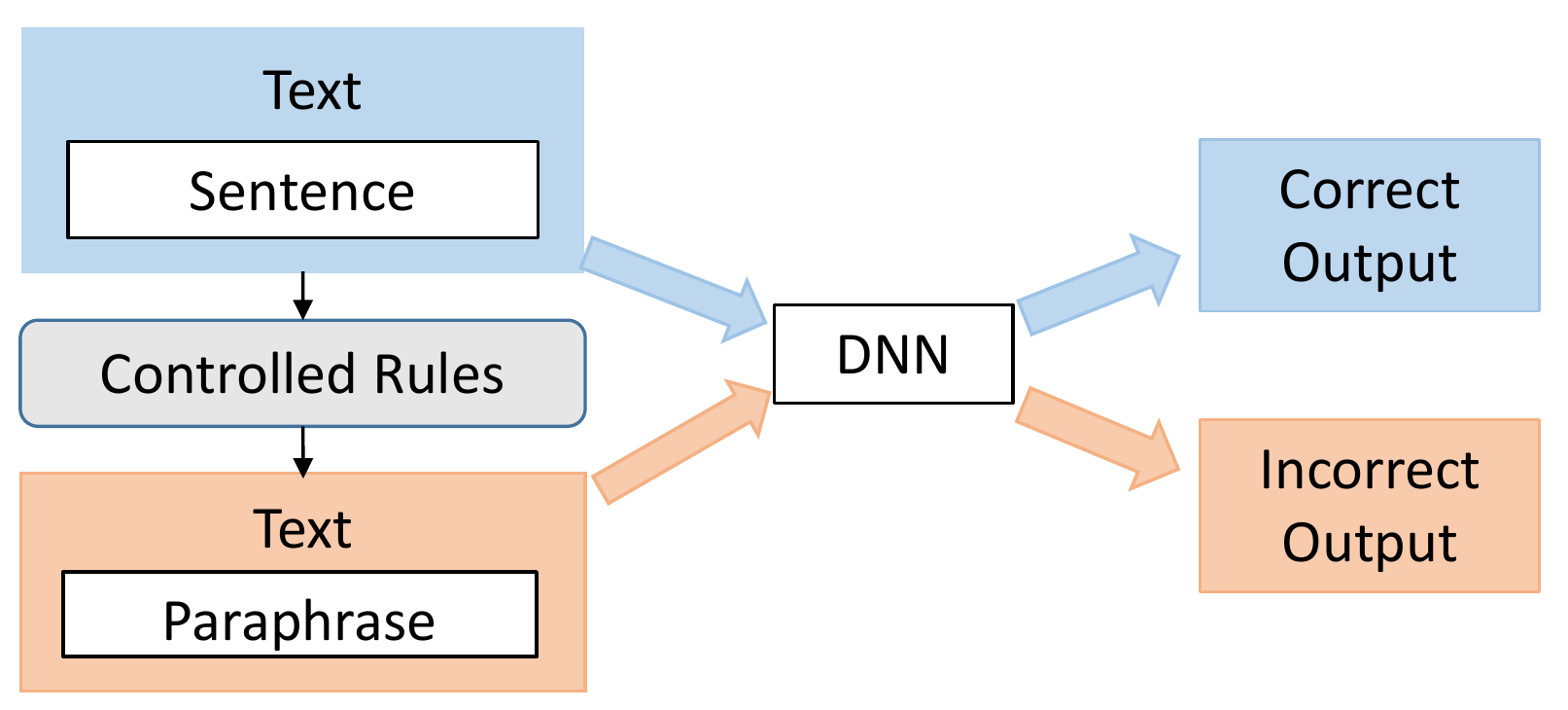}
\caption{General principle of paraphrase-based adversaries. Carefully designed (controlled) paraphrases are regarded as adversarial examples, which fool DNN to produce incorrect output. }
\label{fig:bl-para-wf}
\end{figure}

\subsubsection{Paraphrase-based Adversaries}
\label{sec:subsub_paraphrase}
SCPNs \cite{naacl/IyyerWGZ18} produces a paraphrase of the given sentence with desired syntax by inputting the sentence and a targeted syntactic form into an encoder-decoder architecture. Specifically, the method first encodes the original sentence, then inputs the paraphrases generated by back-translation and the targeted syntactic tree into the decoder, whose output is the targeted paraphrase of the original sentence. One major contribution lies on the selection and processing of the parse templates. The authors trained a parse generator separately from SCPNs and selected 20 most frequent templates in PARANMT-50M. After generating paraphrases using the selected parse templates, they further pruned non-sensible sentences by checking n-gram overlap and paraphrastic similarity.  The attacked classifier can correctly predict the label of the original sentence but fails on its paraphrase, which is regarded as the adversarial example. SCPNs had been evaluated on sentiment analysis and textual entailment DNNs and showed significant impact on the attacked models. Although this method use target strategy to generate adversarial examples, it does not specify targeted output. Therefore, we group it to untarget attack. 
% ah: 
Furthermore, the work in \cite{acl/SinghGR18} used the idea of paraphrase generation techniques that create semantically equivalent adversaries (SEA). They  generated paraphrases of a input sentence $x$, and got predictions from $f$ until the original prediction is changed with considering the semantically equivalent to $x'$  that is $1$ if $x$ is semantically equivalent to $x'$ and $0$ otherwise as shown in Eq.(\ref{eq:SEA}). After that, this work proposes semantic-equivalent rule based method for generalizing these generated adversaries into semantically equivalent rules in order to understand and fix the most impactful bug.
\begin{eqnarray}
\label{eq:SEA}
&\mathbf{SEA(x,x')} = \mathbf{1} [SemEq(x,x') \wedge f(\mathbf{x}) \neq  f(\mathbf{x}')]
\end{eqnarray}

 \subsubsection{GAN-based Adversaries}
 %
%GAN like
Some works proposed to leverage Generative Adversarial Network (GAN) \cite{nips/GoodfellowPMXWOCB14} to generate adversaries \cite{zhao2017generating}. 
The purpose of adopting GAN is to make the adversarial examples mroe natural. 
In \cite{zhao2017generating},  the model proposed to generate adversarial exsamples consists of two key components: a GAN, which generate fake data samples, and an inverter that maps input $x$ to its latent representation $z'$). 
The two components are trained on the original input by minimizing  reconstruction error between original input and the adversarial examples.
Perturbation is performed in the latent dense space by identifying the perturbed sample $\hat{z}$ in the neighborhood  of $z'$. Two search approaches, namely \textit{iterative stochastic search} and \textit{hybrid shrinking search}, are proposed to identify the proper $\hat{z}$. 
However, it requires querying the attacked model each time to find the $\hat{z}$ that can make the model give incorrect prediction. Therefore, this method is quite time-consuming. The work is applicable to both image and textual data as it intrinsically eliminates the problem raised by the discrete attribute of textual data.   The authors evaluated their method on three applications namely:  textual entailment, machine translation and image classification.

%

%  \subsubsection{Probabilistic-based Adversaries}
% %

% \wei{need to refine or delete this paragraph} 
% The work \cite{yang2018greedy} proposed a two-stage probabilistic framework to generate adversarial examples with discrete input (text). 
% In the first stage key features to be perturbed are identified, while in the second stage, perturbation is performed by values chosen from a pre-fixed dictionary. 
% Two attacks based on this framework are proposed: 
% Greedy attack evaluates models with single-feature perturbed inputs. 
% Gumbel attack leverages the Gumbel trick \cite{iclr17/maddisonconcrete,iclr17/jangcategorical} to learn a parametric sampling distribution for perturbation.

\subsubsection{Substitution}
 %
%
% Malware
The work in \cite{hu2017black} proposes a \cll{black-box framework that attacks RNN model for malware detection}. The framework consists of two models: one is a generative RNN, the other is a substitute RNN. The generative RNN aims to generate adversarial API sequence from the malware's API sequence. It is based on the \cll{seq2seq model proposed in~\cite{nips/SutskeverVL14}}. It particularly generates a small piece of API sequence and inserts the sequence after the input sequence.
The substitute RNN, which is a bi-directional RNN with attention mechanism, is to  mimic the behavior of the attacked RNN. Therefore, generating adversarial examples will not query the original attacked RNN, but its substitution. The substitute RNN is trained on both malware and benign sequences, as well as the Gumbel-Softmax outputs of the generative RNN. \cll{Here, Gumbel-softmax is used to enable the joint training of the two RNN models, because the original output of the generative RNN is discrete}. \cll{Specifically, it enables the gradient to be back-propagated from generative RNN to substitute RNN}. 
This method performs attack on API, which is represented as a one-hot vector, i.e., given $M$ APIs, the vector for the $i$-th API is an M-dimensional binary vector that the $i$-th dimension is 1 while other dimensions are 0s.

\begin{table}[t]\footnotesize
\begin{tabular}{|p{19mm}|p{7mm}|c|c|p{33mm}|p{15mm}|p{11mm}|}
\hline
\textbf{Strategy}  & \textbf{Work}  & \textbf{Granularity} & \textbf{Target}  & \textbf{Attacked Models}  &  \textbf{Perturb Ctrl.}   &  \textbf{App.}    \\ \hline \hline
\multirow{2}{*}{Concatenation}    & \cite{emnlp/JiaL17}          & word      & N  & BiDAF, Match-LSTM &  --  & MRC  \\  
\cline{2-7}  & \cite{naacl/WangM_naacl18}  & word, character   & N &  BiDAF+Self-Attn+ELMo \cite{naacl/PetersNIGCLZ18}  &  -- &  MRC \\
 \cline{2-7}  & \cite{conll/BlohmJSYV18}  & word, sentence &  N & \cite{wang2016compare,corr/abs-1709-05036,dzendzik2017framed}, CNN, LSTM and  ensembles  &  Number of changes  &  MRC, QA \\   \hline
 \multirow{7}{*}{Edit}  & \cite{belinkov2017synthetic} & character, word   & N & Nematus \cite{eacl/SennrichFCBHHJL17}, char2char \cite{tacl/LeeCH17}, charCNN \cite{aaai/KimJSR16} &  --  & MT \\ 
  \cline{2-7} & \cite{conll/NiuB18}     & word, phrase    & N &  VHRED \cite{aaai/SerbanSLCPCB17}+attn, RL in \cite{emnlp/LiMRJGG16}, DynoNet \cite{acl/HeBEL17} & --  & DA \\ 
  \cline{2-7} & \cite{gao2018black}  & character, word  & N  & Word-level LSTM, Character-level CNN  & --  & SA, TC \\
  \cline{2-7} &\cite{ndss/LiJDLW19} & character, word     & N  & Word-level LSTM, Character-level CNN  & EdDist, JSC, EuDistV, CSE  & SA \\
  \cline{2-7} & \cite{conll/Minervini018}   & word, phrase    & N  & cBiLSTM, DAM, ESIM & Perplexity & NLI \\  
  \cline{2-7} & \cite{emnlp/AlzantotSEHSC18} & word & N & LSTM & EuDistV & SA, TE \\ 
   \cline{2-7} & \cite{corr/ChanMXXLO2018} & word, sentence & N & DNC & -- & QA \\
  \hline
\multirow{2}{*}{Paraphrase-based} & \cite{naacl/IyyerWGZ18} & word  & N  & LSTM & Syntax-ctrl paraphrase & SA and TE \\ 
  \cline{2-7} & \cite{acl/SinghGR18}   & word     & N  & BiDAF, Visual7W \cite{cvpr/ZhuGBF16}, fastText \cite{eacl/GraveMJB17} & Self-defined semantic-equivalency & MRC, SA, VQA \\  
\hline
%Probabilistic-based               & \cite{yang2018greedy}        & character, word    & untargeted \\ \hline
%
\multirow{1}{*}{GAN-based} & \cite{zhao2017generating} & word  & N  & LSTM, TreeLSTM, Google Translate (En-to-Ge) & GAN-constraints  & TE, MT \\ \hline
Substitution  & \cite{hu2017black}     & API   & N  & LSTM, BiLSTM and variants & -- & MD \\  \hline
Reprogramming   & \cite{corr/NeekharaHDK2018}          &  word &  N & CNN, LSTM, Bi-LSTM  &  -- &  TC \\  \hline
\end{tabular}
%\vspace{1mm}
\caption{Summary of reviewed black-box attack methods. MRC: machine reading comprehension; QA: question answering; VQA: visual question answering; DA: dialogue generation; TC: text classification; MT: machine translation; SA: sentiment analysis; NLI: natural language inference; TE: textual entailment; MD: malware detection.
EdDist: edit distance of text, JSC: Jaccard similarity coeffcient, EuDistV: Euclidean distance on word vector, CSE: cosine similarity on word embedding. '-': not available.}
\label{tbl:black}
\end{table}

\subsubsection{Reprogramming }
As aforementioned,  \cite{corr/NeekharaHDK2018} provides both white-box and black-box attacks. We describe black-box attack here. In black-box attack, the authors fomulated the sequence generation as a reinforcement learning problem, and the adversarial reprogramming function $g_\theta$ is  the policy network. Then they applied  REINFORCE-based optimisation to train  $g_\theta$.

\vspace{2mm}
\noindent \textbf{Summary of Black-box Attack.}
We summarise the reviewed black-box attack works in Table \ref{tbl:black}. We highlight four aspects include granularity-on which level the attack is performed; target-whether the method is  target or un-target; the attacked model, perturbation control, and applications.

\subsection{Multi-modal Attacks}
\label{sec:sub_multimodal}
Some works attack DNNs that are dealing with cross-modal data. For example, \cll{the neural models contain an internal component that performs image-to-text or speech-to-text conversion}. Although these attacks are not for pure textual data, we briefly introduce the representative ones for the purpose of a comprehensive review. 

\subsubsection{Image-to-Text}
\cll{Image-to-text models is a class of techniques that generate textual description for an image based on the semantic content of the latter}.

\vspace{1mm}
\noindent \textbf{Optical Character Recognition (OCR)}. 
Recognizing characters from images is a problem named Optical Character Recognition (OCR). 
OCR is a multimodal learning task that takes an image as input and output the recognized text.  
Authors in~\cite{corr/songS_ocr18} proposed a white-box attack on OCR and follow-up NLP applications. They firstly used the original text to render a clean image (conversion DNNs). Then they found words in the text that have antonyms in WordNet and satisfy edit distance threshold. Only the antonyms that are valid and keep semantic inconsistencies will be kept. Later, the method locates the lines in the clean image containing the aforementioned words, which can be replaced by their selected antonyms.   The method then transforms the target word to target sequence. Given the input/target images and sequences, the authors formed the generating of adversarial example is an optimisation problem: 
\begin{eqnarray}
\label{eq:ocr}
& \min_{\omega} c\cdot J_{CTC}f(\mathbf{x'},t')+||\mathbf{x}-\mathbf{x}'||_2^2  \\
& \mathbf{x'}= (\alpha\cdot \tanh({\omega})+\beta)/2  \label{eq:xp}  \\
&\alpha = (\mathbf{x}_{max}-\mathbf{x}_{min})/2, \beta = (\mathbf{x}_{max}+\mathbf{x}_{min})/2 \nonumber \\
&  J_{CTC}(f(\mathbf{x},t))=-\log p(t|\mathbf{x}) \label{eq:lctc}
\end{eqnarray}
where $f(\mathbf{x})$ is the neural system model, $J_{CTC}(\cdot)$ is the  Connectionist Temporal Classification (CTC) loss function, $\mathbf{x}$ is the input image, $t$ is the ground truth sequence, $\mathbf{x}'$ is the adversarial example, $t'$ is the target sequence, $\omega,\alpha, \beta$ are parameters  controlling adversarial examples to satisfy the box-constraint of $\mathbf{x}' \in [\mathbf{x}_{min},\mathbf{x}_{max}]^p$, where $p$ is the number of pixels ensuring valid $x'$. 
After generating adversarial examples, the method replaces the images of the corresponding lines in the text image. 
The authors evaluated this method in three aspects: single word recognition, whole document recognition, and NLP applications which based on the recognised text (sentiment analysis and document categorisation specifically). They also addressed that the proposed method suffers from limitatios such as low transferability across data and models, and physical unrelalizability.

\vspace{1mm}
\noindent \textbf{Scene Text Recognition (STR)}.  STR is also \cll{an} image-to-text application. In STR, the entire image is mapped to word strings directly. \cll{In contrast, the recognition in OCR is a pipeline process: first segments the words to characters, then performs the recognition on single characters.}
\cll{AdaptiveAttack \cite{corr/YuanHL2018} evaluated the possibility of performing adversarial attack for scene text recognition}. The authors proposed two attacks, namely basic attack and adaptive attack. Basic attack is similar to the work in~\cite{corr/songS_ocr18} and it also formulates the adversarial example generation as an optimisation problem:  
\begin{eqnarray}
\label{eq:str}
& \min_{\omega} J_{CTC}f(\mathbf{x'},t')+ \lambda\mathcal{D}(\mathbf{x},\mathbf{x}') \\
& \mathbf{x'}= \tanh({\omega}) \label{eq:xp2}  
\end{eqnarray}
where $\mathcal{D}(\cdot)$ is Euclidean distance. 
The differences to \cite{corr/songS_ocr18}  lie on the definition of $\mathbf{x'}$ (Eq. (\ref{eq:xp}) vs Eq. (\ref{eq:xp2})), and the distance measurement between $\mathbf{x}$,  $\mathbf{x'}$ ($L_2$ norm vs Euclidean distance), and the parameter $\lambda$, which balances the importance of being adversarial example and close to the original image.  As searching for proper $\lambda$ is quite time-consuming, the authors proposed another method to adaptively find $\lambda$. They named this method Adaptive Attack, in which they defined the likelihood of a sequential classification task following a Gaussian  distribution and derived the adaptive optimization for sequential adversarial examples as:
\begin{eqnarray}
\label{eq:str-sq}
& \min \frac{||\mathbf{x}-\mathbf{x}'||_2^2}{\lambda_1^2}  + \frac{J_{CTC}f(\mathbf{x'},t')}{\lambda_2^2} +\log\lambda_1^2+T\log\lambda_2^2+\frac{1}{\lambda_2^2} 
\end{eqnarray}
where $\lambda_1$ and $\lambda_2$ are two parameters to balance perturbation and CTC loss, $T$ is the number of valid paths given targeted sequential output. 
 Adaptive Attack can be applied to generate  adversarial examples on both non-sequential and sequential classification problems. Here we only highlight the equation for sequential data. 
The authors evaluated their proposed methods on  tasks that targeting the text insertion, deletion and substitution in output. The results demonstrated that Adaptive Attack is much faster than basic attack. 

\vspace{1mm}
\noindent \textbf{Image Captioning}. Image captioning is another multimodal learning task that takes an image as input and generates a textual caption describing its visual contents.
Show-and-Fool \cite{acl/chenVisualImageCap} generates adversarial examples to attack the CNN-RNN based image captioning model. The CNN-RNN model attacked uses a CNN as encoder for image feature extraction and a RNN as decoder for caption generation.  
Show-and-Fool has two attack strategies:  \textit{targeted caption} (i.e., the generated caption matches the target caption) and \textit{targeted keywords} (i.e., the generated caption contains the targeted keywords). In general, they formulated the two tasks using the following formulation: 
\begin{eqnarray}
\label{eq:ic}
& \min_\omega c\cdot J(\mathbf{x'}) + ||\mathbf{x'}-\mathbf{x}||_2^2 \\
& \mathbf{x'}=\mathbf{x}+\eta \nonumber  \\
& x=\tanh{(y), ~~ \mathbf{x'}=\tanh{(\omega+y)}} \nonumber 
\end{eqnarray}
where $c > 0$ is a pre-specified regularization constant, $\eta$ is the perturbation, $\omega, y$ are parameters controlling $\mathbf{x'}\in [-1,1]$. The difference between these two strategies is the definition of the loss function $J(\cdot)$. 
For targeted caption strategy, provided the targeted caption as $S=(S_1, S_2, ... S_t,...S_N)$, where $S_t$ refers to the index of the $t$-th word in the vocabulary and $N$ is the length of the caption, the loss is formulated as:
\begin{eqnarray}
\label{eq:ic1}
& J_{S,logit}(\mathbf{x'}) =  \sum_{t=2}^{N-1}\max\{-\epsilon, \max_{k\neq S_t}\{z_t^{(k)}\}-z_t^{(S_t)}\} 
\end{eqnarray}
where $S_t$ is the target word, $z_t^{(S_t)}$ is the \textit{logit} of the target word.  In fact, this method mininises the difference between the maximumn \textit{logit} except $S_t$, and the logit of $S_t$. 
For the targeted keywords strategy, given the targeted keywords $\mathcal{K}:={K_1,...,K_M}$, the loss function is:
\begin{eqnarray}
\label{eq:ic2}
& J_{K,logit}(\mathbf{x'}) =   \sum_{j=1}^{M}\min_{t\in[N]}\{\max\{-\epsilon, \max_{k\neq K_j}\{z_t^{(k)}\}-z_t^{(K_j)}\}\} 
\end{eqnarray}
The authors performed extensive experiments on Show-and-Tell \cite{cvpr/VinyalsTBE15} and varied the parameters in the attacking loss. They found that Show-and-Fool is not only effective on attacking Show-and-Tell, the CNN-RNN based image captioning model, but is also highly transferable to another model Show-Attend-and-Tell \cite{icml/XuBKCCSZB15}.

\vspace{1mm}
\noindent \textbf{Visual Question Answering (VQA)}. 
Given an image and a natural language question about the image, VQA is to provide an accurate answer in natural language. 
The work in \cite{cvpr/XuCLRDS18} proposed a iterative optimisation method to attack two VQA models. %All of the attacked models contain language generation component, localization and attention mechanism. 
The objective function proposed maximises the probability of the target answer and \cll{unweights} the  preference of adversarial examples with smaller distance to the original image when this distance is below a threshold. Specifically, the objective  contains three components. The first one is similar to Eq. (\ref{eq:str}), that replaces the loss function to the loss of the VQA model and using $||\mathbf{x}-\mathbf{x'}||_2/\sqrt{N}$ as distance between $\mathbf{x'}$ and $\mathbf{x}$.  The second component maximises the difference between the softmax output and the prediction when it is different with the target answer. The third component ensures the distance between $\mathbf{x'}$ and $\mathbf{x}$ is under a lower bound. 
\cll{The attacks are evaluated by checking whether better success rate is obtained over the previous attacks,} and the confidence score of the model to predict the target answer. 
Based on the evaluations, the authors concluded that \st{that} attention, bounding box localization and compositional internal structures are vulnerable to adversarial attacks. 
This work also attacked a image captioning neural model. We refer the original paper for further information. 

%\cll{The work in \cite{cvpr/XuCLRDS18} proposed a method to attack one image captioning neural model and two visual question answering (VQA) models. All of the attacked models contain language generation component, localization and attention mechanism. The image captioning model DenseCap utilizes a localization network (RNN) to predict regions and a CNN+RNN module to generate textual output from images}. This work attacked the CNN+RNN module in DenseCap by using targeted region-caption pairs as groundtruth and optimizing the DenseCap loss as well as the distance between perturbed and original images on the adversarial examples. For the VQA attack, the work proposed a loss function that \cll{maximizes} the probability of the target answer and \cll{unweights} the preference of adversarial examples with smaller distance to the original image when this distance is below a threshold. \cll{The attacks are evaluated by checking whether better success rate is obtained over the previous attacks}. 

\begin{table}\footnotesize
\begin{tabular}{|l|l|c|c|c|p{22mm}|p{12mm}|}
\hline
\textbf{Multi-modal}        & \textbf{Application}  & \textbf{Work} & \textbf{Target} & \textbf{Access} & \textbf{Attacked Models}  & \textbf{Perturb Ctrl.}   \\ \hline \hline
\multirow{4}{*}{Image-to-Text} & Optical Character Recognition  & \cite{corr/songS_ocr18}  & Y  & white-box & Tesseract \cite{tesseract} & $L_2$, EdDist\\ 
\cline{2-7}  & Scene Text Recognition & \cite{corr/YuanHL2018}  & Y & white-box & CRNN \cite{shi2017end} & $L_2$\\  
\cline{2-7}  & Image Captioning   & \cite{acl/chenVisualImageCap} & Y & white-box & Show-and-Tell \cite{cvpr/VinyalsTBE15}  & $L_2$\\ 
\cline{2-7}  & Visual Question Answering   & \cite{cvpr/XuCLRDS18}  & Y  & white-box & MCB \cite{emnlp/FukuiPYRDR16}, N2NMN \cite{iccv/HuARDS17} &  $L_2$ \\ 
\cline{2-7} & Visual-Semantic Embeddings  & \cite{coling/ShiMXJS18} & N & black-box & VSE++ \cite{corr/FaghriFKF17}  & --\\ \hline 
Speech-to-Text  & Speech  Recognition  & \cite{sp/Carlini018}    & Y  & white-box &  DeepSpeech \cite{corr/HannunCCCDEPSSCN14}  & $L_2$ \\ \hline
\end{tabular}
\label{tbl:cross}
\caption{Summary of reviewed cross-modal attacks. EdDist: edit distance of text, -: not available. } 
\end{table}

\vspace{1mm}
\noindent \textbf{Visual-Semantic Embeddings (VSE)}. 
The aim of VSE is to bridge natural language and the underlying visual world. In VSE, the embedding spaces of both images and descriptive texts (captions) are jointly optimized and aligned.  \cite{coling/ShiMXJS18} attacked the latest VSE model by generating adversarial examples in the test set and evaluated the robustness of the VSE modesls. They performed the attack on textual part by introducing three method: i) replace nouns in the image captions utilizing the hypernymy/hyponymy relations in WordNet; ii) change the numerals to different ones and singularize or pluralize the corresponding nouns when necessary; iii) \cll{detect the relations  and shuffle the non-interchangeable noun phrases or replace the prepositions. This  method can be considered as a black-box edit adversary }.

\subsubsection{Speech-to-Text}
% speech to text
\cll{Speech-to-text is also known as speech recognition. The task is to recognize and translate the spoken language into text automatically}. 
\cll{\cite{sp/Carlini018} attacked a state-of-the-art speech-to-text transcription neural network (based on LSTM), named DeepSpeech}. 
Given a natural waveform, the authors constructed a audio perturbation that is almost inaudible but can be recognized by adding into the original waveform. The perturbation is constructed by adopting the idea from C\&W method (refers to section \ref{sec_subsub:cw} ), which measures the image distortion by the maximum amount of changed pixels. Adapting this idea, \cll{they measured the audio distortion by calculating relative loudness of an audio and proposed to use Connectionist Temporal Classification loss for the optimization task}. Then they solved this task with Adam optimizer \cite{iclr/KingmaB15}.

\subsection{Benchmark Datasets by Applications}
\label{sec:data}
In recent years, neural networks gain success in different NLP domains and the popular applications include text classification, reading comprehension, machine translation, text summarization, question answering, dialogue generation, to name a few. 
%
%NN models have gain success in different NLP domains. Some example works are for i) text classification, including text categorization \cite{emnlp/Kim14} and sentiment analysis\cite{coling/SantosG14}; ii) reading comprehension \cite{corr/SeoKFH16}; iii) paraphrase detection \cite{nips/HuLLC14}; iv) machine translation \cite{emnlp/ChoMGBBSB14}; v) document summarization \cite{corr/DenilDKBF14}; vi) question answering \cite{acl/YihCHG15}; vii) named entity recognition \cite{corr/LampleBSKD16}, to name a few. 
%%
%However, not all NLP applications receive the research attention on the attack to their neural network models. 
%
In this section, we review the current works on generating adversarial examples on the neural networks in the perspective of NLP applications. Table \ref{tbl:data} summarizes the works we reviewed in this article according to their application \cll{domain}. We further list the benchmark datasets used in these works in the table as auxiliary information-- thus we refer readers to the links/references we collect for the detailed descriptions of the datasets. Note that the auxiliary datasets which help to generate adversarial examples are not included. \cll{Instead, we only present the dataset used to evaluate the attacked neural networks}.

\begin{table*}\footnotesize
\begin{tabular}{|c|l|p{30mm}|p{45mm}|}
\hline
\multicolumn{2}{|l|}{\bf Applications}                             & \bf Representative Works & \bf Benchmark Datasets \\ \hline \hline
\multirow{8}{*}{Classification}      & Text Classification     &  \cite{liang2017deep,gong2018adversarial,gao2018black,acl/EbrahimiRLD18,sato2018interpretable,corr/NeekharaHDK2018}   &  DBpedia, Reuters Newswires, AG's news, Sogou News, Yahoo! Answers, RCV1, Surname Classification Dataset \\ 
\cline{2-4} & Sentiment Analysis      &  \cite{papernot2016crafting,ecir/SamantaAdversarial,gao2018black,acl/EbrahimiRLD18,sato2018interpretable,corr/NeekharaHDK2018,naacl/IyyerWGZ18,acl/SinghGR18} 
 &  SST,  IMDB Review, Yelp Review, Elec, Rotten Tomatoes Review, Amazon Review, Arabic Tweets Sentiment                 \\ 
 \cline{2-4}  & Spam Detection    & \cite{gao2018black} &    Enron Spam, Datasets from \cite{nips/ZhangZL15}         \\ 
\cline{2-4} & Gender Identification   & \cite{ecir/SamantaAdversarial}&     Twitter Gender     \\ \cline{2-4} & Grammar Error Detection &  \cite{sato2018interpretable} &     FCE-public       \\ \cline{2-4}& Medical Status Prediction &  \cite{kdd/SunTYWZ18} &  Electronic Health Records (EHR)   \\ \cline{2-4}  & Malware Detection &  \cite{grosse2016adversarial,ESORICS17/GrosseAdvMalware,rosenberg2017generic,hu2017black,sp/Al-DujailiHHO18}  &  DREBIN, Microsoft Kaggle                   \\ 
\cline{2-4} & Relation Extraction    & \cite{wu2017adversarial,emnlp/BekoulisDDD18} & NYT Relation, UW Relation,  ACE04,  CoNLL04 EC,  Dutch Real Estate Classifieds, Adverse Drug Events \\  
\hline
\multicolumn{2}{|l|}{Machine Translation}&  \cite{belinkov2017synthetic,coling/EbrahimiLD18,cheng2018seq2sick,zhao2017generating}   &   TED Talks, WMT'16 Multimodal Translation Task    \\ \hline 
\multicolumn{2}{|l|}{Machine Comprehension} & \cite{emnlp/JiaL17,naacl/WangM_naacl18,conll/BlohmJSYV18,corr/ChanMXXLO2018} &  SQuAD, MovieQA Multiple Choice, Logical QA \\ \hline 

\multicolumn{2}{|l|}{Text Summarization}&  \cite{cheng2018seq2sick}     &   DUC2003, DUC2004, Gigaword   \\ \hline 
\multicolumn{2}{|l|}{Text Entailment}  & \cite{acl/KangAdventure,conll/Minervini018,naacl/IyyerWGZ18,zhao2017generating}  &  SNLI, SciTail, MultiNLI, SICK \\ \hline 
\multicolumn{2}{|l|}{POS Tagging}&   \cite{naacl/YasunagaKR18}    &  WSJ portion of PTB, Treebanks in UD                  \\ \hline  
\multicolumn{2}{|l|}{Dialogue System}&  \cite{conll/NiuB18}   &  Ubuntu Dialogue, CoCoA,                  \\ \hline  \hline %Switchboard Dialogue Act, OpenSubtitles 

\multirow{4}{*}{Cross-model}      & Optical Character Recognition     &  \cite{corr/songS_ocr18}   &  Hillary Clinton's emails \\ \cline{2-4}
                                     & Scene Text Recognition      &  \cite{corr/YuanHL2018} 
 &  Street View Text, ICDAR 2013, IIIT5K               \\ \cline{2-4} 
                                     & Image Captioning   & \cite{acl/chenVisualImageCap,cvpr/XuCLRDS18} &    MSCOCO,  Visual Genome      \\ \cline{2-4} 
                                     & Visual Question Answering   & \cite{cvpr/XuCLRDS18}&     Datasets from \cite{iccv/AntolALMBZP15}, Datasets from \cite{cvpr/ZhuGBF16}             \\ \cline{2-4} 
                                     & Visual-Semantic  Embedding & \cite{coling/ShiMXJS18}  & MSCOCO \\ \cline{2-4}
                                     & Speech Tecognition &  \cite{sp/Carlini018} &     Mozilla Common Voice             \\ \hline
\end{tabular}
\label{tbl:apps}
\caption{Attacked Applications and Benchmark Datasets}
\label{tbl:data}
\end{table*}

\textbf{Text Classification.} 
\cll{Majority of the surveyed works attack the deep neural networks for text classification, since these tasks can be framed as a classification problem}. 
Sentiment analysis aims to classify the sentiment to 
%\ah{(there are two, three and five groups for sentiment analysis)} three 
several groups (e.g., in 3-group scheme: neural, positive and negative).  Gender identification, Grammatical error detection and malware detection can be framed as binary classification problems. 
%\ah{(Table 4 present Relation Extraction ,spam detection under Text classification. But they are not presented here under which type of classification binary or multi)} 
Relation extraction can  be formulated as single or multi-classification problem. 
Predict medical status is a multi-class problem that the classes are defined by medical experts. 
These works usually use multiple datasets to evaluate their attack strategies to show the generality and robustness of their method.
 \cite{liang2017deep} used DBpedia ontology dataset \cite{semweb/LehmannIJJKMHMK15} to classify the document samples into 14 high-level classes.  
\cite{gong2018adversarial} used IMDB movie reviews \cite{acl/MaasDPHNP11} for sentiment analysis, and Reuters-2 and Reuters-5 newswires dataset provided by NLTK package\footnote{https://www.nltk.org/} for categorization. 
\cite{papernot2016crafting} used a un-specified movie review dataset for sentiment analysis. 
\cite{ecir/SamantaAdversarial} also used IMDB movie review dataset for sentiment analysis. The work also performed gender classification on and Twitter dataset\footnote{https://www.kaggle.com/crowdflower/twitter-user-gender-cla2013.} for gender detection. %~\fixme{the same work with~\cite{ecir/SamantaAdversarial}???}. 
\cite{gao2018black} performed spam detection on Enron Spam Dataset \cite{ceas/MetsisAP06} and adopted six large datasets \cll{from} \cite{nips/ZhangZL15}, i.e., AG's news\footnote{https://www.di.unipi.it/˜gulli/}, 
Sogou news \cite{www/WangZMR08},  DBPedia ontology dataset, Yahoo! Answers\footnote{Yahoo! Answers Comprehensive Questions and Answers
version 1.0 dataset through the Yahoo! Webscope program.} for text categorization and Yelp reviews\footnote{Yelp Dataset Challenge in 2015}, Amazon reviews \cite{recsys/McAuleyL13} for sentiment analysis. 
\cite{acl/EbrahimiRLD18} also used AG's news for text classification. Further, they used Stanford Sentiment Treebank (SST) dataset \cite{emnlp/SocherPWCMNP13} for sentiment analysis. 
\cite{sato2018interpretable} conducted evaluation on three tasks: sentiment analysis (IMDB movie review, Elec \cite{nips/JohnsonZ15}, Rotten Tomatoes \cite{acl/PangL05}), text categorization (DBpedia Ontology dataset and RCV1 \cite{jmlr/LewisYRL04}) and grammatical error detection (FCE-public \cite{acl/YannakoudakisBM11}). 
%
%\cite{kdd/LiY18}  targeted two applications: text categorization (AG's news corpus, DBPedia ontology dataset) and sentiment analysis (IMDB movie review, Yelp review). It is not adversarial attack.  
%
%\cite{yang2018greedy} used IMDB movie review for sentiment analysis, and  AG's news and Yahoo! Answers for text categorization. 
% 
%\cite{wong2017dancin} used Enron Spam dataset for spam detection. 
%
\cite{kdd/SunTYWZ18} generated adversarial examples on the neural medical status prediction system \cll{with} real-world electronic health records data. 
Many works target the malware detection models. 
\cite{grosse2016adversarial,ESORICS17/GrosseAdvMalware} performed attack on neural malware detection systems. They used DREBIN dataset which contains both benigh and malicious android applications \cite{ndss/ArpSHGR14}. 
\cite{rosenberg2017generic} collected benigh windows application files and used Microsoft Malware Classification Challenge dataset \cite{corr/RonenRFTA2018} as the malicious part. 
\cite{hu2017black} crawled 180 programs with corresponding behavior reports from a website for malware analysis\footnote{https://malwr.com/}. 
70\% of the crawled programs are malware. 
%
%
%\cite{sp/Al-DujailiHHO18} created the malicious and benign Portable Executable (PE) files from VirusShare\footnote{https://virusshare.com/}. 
%
\cite{corr/NeekharaHDK2018} proposed another kind of attack, called reprogramming. They specifically targeted the text classification neural models and used four datasets to evaluate their attack methods: Surname Classification Dataset\footnote{Classifying names with a character-level rnn - pytroch tutorial. % \url{https: //pytorch.org/tutorials/intermediate/char_rnn_classification_tutorial.html}.
}, Experimental Data for Question Classification \cite{coling/LiR02}, Arabic Tweets Sentiment Classification Dataset \cite{abdulla2013arabic} and IMDB movie review dataset. 
%
%Relation Extraction.
%
In \cite{wu2017adversarial}, the authors modelled the relation extraction as a classification problem, where the goal is to predict the relations exist between entity pairs given text mentions.  They used two relation datasets:  NYT dataset \cite{pkdd/RiedelYM10} and UW dataset \cite{naacl/LiuSBLLW16}.
%
%Joint Extraction.
\cll{The work \cite{emnlp/BekoulisDDD18} targeted at improving the efficacy of the neural networks for joint entity and relation extraction.} %More precisely, the authors proposed to include adversarial training in the model optimization process. 
Different to the method in \cite{wu2017adversarial}, the authors modelled  the relation extraction task as a multi-label head selection problem. 
%They also adopted the adversarial training method in \cite{miyato2016adversarial}.
\cll{The four datasets are used in their work}: ACE04 dataset \cite{L04-1011}, CoNLL04 EC tasks \cite{conll/RothY04}, Dutch Real Estate Classifieds (DREC) dataset \cite{eswa/BekoulisDDD18}, and Adverse Drug Events (ADE) \cite{jbi/GurulingappaRRFHT12}.

\textbf{Machine Translation.}  
Machine Translation works on parallel datasets, one of which uses source language and the other one is in the \cll{target} language.  
\cite{belinkov2017synthetic} used the TED talks parallel corpus prepared for IWSLT 2016 \cite{mauro2012wit3} for testing the NMT systems. They also collected French, German and Czech corpus for generating natural noises to build a look-up table which contains possible lexical replacements that later be used for generating adversarial examples. 
\cite{coling/EbrahimiLD18} also used the same TED talks corpus and used German to English, Czech to English, and French to English pairs. 

\textbf{Machine Comprehension.}
Machine comprehension datasets usually provide context documents or paragraphs to the machines. Based on the comprehension of the contexts,  machine comprehension models can answer a question.  
Jia and Liang are one of the first to consider the textual adversary and they targeted the neural machine comprehension models \cite{emnlp/JiaL17}.  They used the  Stanford Question Answering Dataset (SQuAD) to evaluate the impact of their attack on the neural machine comprehension models. SQuAD is a widely recognised benchmark dataset for machine comprehension. 
\cite{naacl/WangM_naacl18} followed the previous works and also worked on SQuAD dataset. 
%
%\cite{acl/WallaceB18} evaluated their attacks on 2017 NIPS Human-Computer Question Answering competition \cite{Springer/BGFR18}. 
%
Althouth the focus of the work \cite{conll/BlohmJSYV18} is to develop a robust machine comprehension model rather than attacking MC models, they used the adversarial examples to evaluate their proposed system. They used MovieQA multiple choice question answering dataset \cite{cvpr/TapaswiZSTUF16} for the evaluation. 
\cite{corr/ChanMXXLO2018} targeted attacks on differentiable neural computer (DNC), which is a novel computing machine with DNN. They evaluated the attacks on logical question answering using bAbI tasks\footnote{https://research.fb.com/downloads/babi/}.% \ah{ ( I am not sure why you put this work \cite{corr/ChanMXXLO2018} in Machine Translation, because  The bAbI dataset has question & answer tasks)} 

\textbf{Text Summarization.} 
The goal for text summarization is to summarize the core meaning of a given document or paragraph with succinct expressions. 
There is no surveyed papers that only target the application of text summarization. 
\cite{cheng2018seq2sick} evaluated their attack on multiple applications including text summarization and they used  DUC2003\footnote{http://duc.nist.gov/duc2003/tasks.html}, DUC2004\footnote{http://duc.nist.gov/duc2004/}, and Gigaword\footnote{https://catalog.ldc.upenn.edu/LDC2003T05} for evaluating the effectiveness of adversarial examples. 

\textbf{Text Entailment.}  
The fundamental task of text entailment is to decide whether a premise text entails a hypothesis, i.e., the truth of one text fragment follows from another text. 
\cite{acl/KangAdventure} assessed various models on two entailment datasets: Standord Natural Lauguage Inference (SNLI) \cite{emnlp/BowmanAPM15} and SciTail \cite{aaai/KhotSC18}.
\cite{conll/Minervini018} also used SNLI dataset. Furthermore, they used MultiNLI \cite{naacl/WilliamsNB18} dataset. 

\textbf{Part-of-Speech (POS) Tagging.} 
The purpose for POS tagging is to resolve the part-of-speech for each word in a sentence, such as noun, verb etc. It is one of the fundamental NLP tasks to facilitate other NLP tasks, e.g., syntactic parsing. Neural networks are also adopted for this NLP task. \cite{naacl/YasunagaKR18} adopted the method in \cite{miyato2016adversarial} to build a more robust neural network by introducing adversarial training, but they applied the strategy (with minor modifications) in  POS tagging.  By training on the mixture of clean and adversarial example, the authors found that adversarial examples not only help improving the  tagging accuracy, but also contribute to downstream task of dependency parsing and is generally effective in different sequence labelling tasks. The datasets used in their evaluation include: the Wall Street Journal (WSJ) portion of the Penn Treebank (PTB) \cite{coling/MarcusSM94} and  treebanks from
Universal Dependencies (UD) v1.2 \cite{nivre2015universal}.

\textbf{Dialogue Generation.} 
Dialogue generation is a fundamental component for real-world virtual assistants such as Siri\footnote{https://www.apple.com/au/siri/} and Alexa\footnote{https://en.wikipedia.org/wiki/Amazon\_Echo}. \cll{It is the text generation task that automatically generate a response given a post by the user}. 
\cite{conll/NiuB18} is one of the first to attack the generative dialogue models. They used the Ubuntu Dialogue Corpus \cite{sigdial/LowePSP15} and Dynamic Knowledge Graph Network with the Collaborative Communicating Agents (CoCoA) dataset \cite{acl/HeBEL17} for the evaluation of their two attack strategies. 
%
%\cite{corr/HeC2018} also used Ubuntu Dialogue Corpus. In addition, they also used Switchboard Dialogue Act Corpus\footnote{http://compprag.christopherpotts.net/swda.html}, which  is a collection of two-sided telephone conversations, annotated with utterance-level dialogue acts. 
%Another dataset, OpenSubtitles data-set\footnote{http://www.opensubtitles.org/} is also used for evaluation. \cll{The} conversations in this dataset contains a large number of egregious sentences. 

\textbf{Cross-model Applications.} 
\cite{corr/songS_ocr18}  evaluated the OCR systems with adversarial examples using  Hillary Clinton's emails\footnote{https://www.
kaggle.com/kaggle/hillary-clinton-emails/data}, which is in the form of images.  They also \cll{conducted} the attack on NLP applications using Rotten Tomatoes and IMDB review datasets.  
The work in \cite{corr/YuanHL2018} attacked the neural networks \cll{designed} for scene text rcognition. They conducted experiments on three standard benchmarks for cropped word image recognition, namely the Street View Text dataset (SVT) \cite{iccv/WangBB11} the ICDAR 
2013 dataset (IC13) \cite{icdar/KaratzasSUIBMMMAH13} and the IIIT 5K-word dataset (IIIT5K) \cite{bmvc/MishraAJ12}. 
\cite{acl/chenVisualImageCap} attacked the image captioning neural models. The dataset they used is the Microsoft COCO (MSCOCO) dataset \cite{eccv/LinMBHPRDZ14}. 
\cite{cvpr/XuCLRDS18} worked on the problems of attacking neural models for image captioning and visual question answering. For the first task, they used Visual Genome dataset \cite{ijcv/KrishnaZGJHKCKL17}.  For the second task, they used the VQA datasets collected and processed in \cite{iccv/AntolALMBZP15}. 
\cite{coling/ShiMXJS18} worked on Visual-Semantic Embedding applications, \cll{where the MSCOCO dataset is used}. 
\cite{sp/Carlini018} targeted the speech recognition problem. The datasets they used is the Mozilla Common Voice dataset\footnote{https://voice.mozilla.org/en}.

\textbf{Multi-Applications}
Some works adapt their attack methods into different applications, namely, they evaluate their method's trasferability across applications.  \cite{cheng2018seq2sick} attacked the sequence-to-sequence models. Specifically, they evaluated their attack on two applications: text summarization and machine translation. For text summarization, as mentioned before, they used three datasets  DUC2003, DUC2004, and Gigaword.  For the machine translation, they \cll{sampled a subset} form WMT'16 Multimodal Translation dataset\footnote{http://www.statmt.org/wmt16/translation-task.html}. 
\cite{naacl/IyyerWGZ18} proposed syntactically adversarial paraphrase and evaluated the attack on sentiment analysis and text entailment applications. They used SST for sentimental analysis and SICK \cite{semeval/MarelliBBBMZ14} for text entailment. 
\cite{zhao2017generating} is a generic approach for generating adversarial examples on neural models. \cll{The applications investigated include} image classification (MINIST digital image dataset), textual entailment (SNLI), and machine translation. 
\cite{miyato2016adversarial} evaluated their attacks on five datasets,\cll{ covering both sentiment analysis (IMDB movie review, Elec product review, Rotten Tomatoes movie review) and text categorization (DBpedia Ontology, RCV1 news articles)}. 
\cite{acl/SinghGR18} targeted two applications. For sentiment analysis, they used Rotten Tomato movie reviews and IMDB movie reviews datasets. 
For visual question answering, they tested on dataset provided by Zhu et al. \cite{cvpr/ZhuGBF16}.

\section{Defense}
\label{sec:defense}

\cll{An essential purpose for generating adversarial examples for neural networks is to utilize these adversarial examples to enhance the model's robustness~\cite{iclr15/goodfellow}}. 
There are two common way\ah{s} in textual DNN to achieve this goal:  \textit{adversarial training} and \textit{knowledge distillation}. 
Adversarial training incorporates adversarial examples in the model training process. 
Knowledge distillation manipulates the neural network model and trains a new model. 
In this section, we introduce some representative studies belonging to these two directions. 
For more comprehensive defense strategies on machine learning and deep leaning models and applications, please refer to  \cite{pr/BiggioR18,access/AkhtarM18}.

\subsection{Adversarial Training}
Szegedy et al. \cite{iclr14/Szegedy} invented  \textit{adversarial training}, a strategy that consists of training a neural network to correctly classify both normal examples and adversarial examples. Goodfellow et  al.~\cite{iclr15/goodfellow} employed explicit training with adversarial examples.  In this section, we describe works utilizing \textit{data augmentation}, \textit{model regularization} and \textit{robust optimization} for the defense purpose on textual adversarial attacks. 

%Some adversarial examples are strong and diverse, so performing  adversarial training cannot necessarily guarantee the improvement of the model performance. \cite{emnlp/AlzantotSEHSC18}}. 

\subsubsection{Data Augmentation} 

\cll{Data augmentation extends the original training set with the generated adversarial examples and try to let the model see more data during the training process}. 
Data augmentation is commonly used against black-box attacks with additional training epochs on the attacked DNN with adversarial examples.

The authors in work \cite{emnlp/JiaL17} try to \cll{enhance} the reading comprehension model with training on the augmented dataset that includes the adversarial examples. \cll{They showed that this data augmentation is effective and robust against the attack that uses the same adversarial examples}. However, their work also demonstrated that this augmentation strategy would be still vulnerable against the attacks with other kinds of adversarial examples.  
\cite{naacl/WangM_naacl18} shared similar idea to augment the training dataset, \cll{but selected further informative adversarial examples as discussed in Section \ref{sec:_subsub_concat}}.

The work in \cite{acl/KangAdventure} trains the text entailment system augmented with adversarial examples. \cll{The purpose} is to make the system more robust. 
\cll{They proposed three methods to generate more data with diverse characteristics:  (1) \textit{knowledge-based}, which replaces words with their hypernym/hyponym provided in several given knowledge bases; (2) \textit{hand-crafted}, which adds negations to the the existing entailment; (3) \textit{neural-based}, which leverages a seq2seq model to generate an entailment examples by enforcing the loss function to measure the cross-entropy between the original hypothesis and the predicted hypothesis.  During the training process, they adopt the idea from generative adversarial network to train a discriminator and a generator, and incorporating the adversarial examples in the discriminator's optimization step}.   

\cite{belinkov2017synthetic} explores another way for data augmentation. It takes  the  average character embedding as a word representation and incorporate it into the input. This approach is intrinsically insensitive to character scrambling such as \textit{swap}, \textit{mid} and \textit{Rand}, thus can resists to   noises caused by these scrambling attacks proposed in the work. However, this defense is ineffective to other attacks that do not perturb on characters' orders. %\ah{(I like if there is answers that explains the reasons why this approach ineffective to other attacks)}  

\subsubsection{Model Regularization}
Model regularization enforces the generated adversarial examples as the regularizer and follows the form of:
\begin{eqnarray}
\label{eq:advtrain}
 \min (J(f(x), y) + \lambda J(f(x'), y)) 
\end{eqnarray}
where $\lambda$ is a hyperparameter.
%  ~\fixme{can not understand the Equations and the description.}

Following~\cite{iclr15/goodfellow}, the work \cite{miyato2016adversarial} constructed the adversarial training with a linear approximation as follows: 
\begin{eqnarray}
\label{eq:advtrainfsgm}
 & -\log p(y|x+-\epsilon g/||g||_2,; \theta) \\
 & g=\partial_x\log p(y|x;\Hat{\theta}) \nonumber 
\end{eqnarray}
where $||g||_2$ is the $L_2$ norm regularization, $\theta$ is the parameter of the neural model, and $\hat{\theta}$ is a constant copy of $\theta$. The difference to  \cite{iclr15/goodfellow} is that, the authors performed the adversarial generation and training in terms of the word embedding.  
Further, they extended their previous work on attacking image deep neural model \cite{iclr16/miyatodistributional}, where \cll{the local distribution smoothness (LDS) is defined as the negative of the KL divergence of two distributions (original data and the adversaries)}.  LDS measures the robustness of the model against the perturbation in local and `virtual' adversarial direction. \cll{In this sense, the adversary is calculated as the direction to which the model distribution is most sensitive in terms of KL divergence}. 
They also \cll{applied this attack strategy on word embedding and performed} adversarial training by adding adversarial examples as regularizer.

The work \cite{sato2018interpretable} follows the idea from  \cite{miyato2016adversarial} and extends the adversarial training on LSTM. The authors followed \cll{FGSM to incorporate} the adversarial training as a regularizer. But in order to enable the interpretability of adversarial examples, i.e., the word embedding of the adversaries \cll{should be valid word embeddings in the vocabulary, they introduced a direction vector which associates the perturbed embedding to the valid word embedding. %This work also propose to incorporate the adversarial training as a semi-supervised learning by adopting method in \cite{miyato2016adversarial}. 
\cite{wu2017adversarial} simply adopts the regularizer utilized in \cite{miyato2016adversarial}, but applies the perturbations on pre-trained word embedding and in  a different task: relation extraction}. 
Other similar works that adopt \cite{miyato2016adversarial}  are  \cite{wu2017adversarial,naacl/YasunagaKR18,sato2018interpretable,emnlp/BekoulisDDD18}. 
 We will not cover all these works in this article, \cll{since} they simply adopting this method. %

\subsubsection{Robust Optimisation}
Madry et al. \cite{iclr/MadryMSTV18} cast DNN model learning as a robust optimization with min-max (saddle point) formulation, which is the composition of an inner non-concave maximization problem (attack)  and an outer non-convex minimization problem (defense). According to  Danskin's theorem,  gradients at inner maximizers correspond to descent directions for the min-max problem, thus the optimization can still apply back-propagation to proceed.  
The approach successfully demonstrated robustness of DNNs against adversarial images by training and learning universally. 
\cite{huang2018adversarial} adopts the idea and applies on malware detection DNN  that handles discrete data.  
Their leaning objective is formulated as:
\begin{eqnarray}
\theta^* = arg \min_{\theta} \mathbb{E}_{(x,y)\sim D} [\max_{x'\in S(x)} L(\theta,x',y) ]
\label{eq:25}
\end{eqnarray}
%
% \begin{eqnarray}
% \theta^* = arg \min_{\theta \in \mathbb{R}^P} \mathbb{E}_{(x,y)\sim D} [\max_{x'\in S(x)} L(\theta,x',y) ]
% \label{eq:25}
% \end{eqnarray}
%
%
where $S(x)$ is the set of binary indicator vectors that preserve the functionality of malware $x$, $L$ is the loss function for the original classification model, $y$ is the \cll{groundtruth} label, $\theta$ is the learnable parameters, $D$ denotes the distribution of data sample $x$. 

It is worth noting that the proposed robust optimisation method is an universal framework under which other adversarial training strategies have natural interpretation. We describe it separately keeping in view its popularity in the literature.

\subsection{Distillation}
Papernot et al. \cite{sp/PapernotM0JS16} proposed distillation as another possible defense against adversarial examples. 
The principle is to use the softmax output (e.g., the class probabilities in classfication DNNs) of the original DNN to train the second DNN, which has the same structure with the original one. 
The softmax of the original DNN is also modified by introducing a \textit{temperature} parameter $T$:
\begin{eqnarray}
q_i = \frac{\exp{(z_i/T)}}{\sum_k \exp{(z_k/T)}}
\label{eq:distill}
\end{eqnarray}
where $z_i$ is input of softmax layer. 
$T$ controls the level of knowledge distillation. When $T=1$, Eq. (\ref{eq:distill}) turns back to the normal softmax function. If $T$ is large, $q_i$ is close to a uniform distribution, when it is small, the function will output more extreme values.   
\cite{ESORICS17/GrosseAdvMalware} adopts distillation defense for DNNs on discrete data and applied a high temperature $T$, as high-temperature softmax is proved to reduce the model sensitivity to small perturbations \cite{sp/PapernotM0JS16}. They trained the second DNN with the augmentation of original dataset and the softmax outputs from the original DNN. From the evaluations, they found  adversarial training is the more effective than distillation. (I like if there is answers that explains why adversarial training is the more effective than distillation )

\section{Discussions and Open Issues} 
\label{sec:future}
Generating textual adversarial examples has relatively shorter history than generating image adversarial examples on DNNs because it is more challenging to make perturbation on discrete data, \cll{and meanwhile preserving  the valid syntactic, grammar and semantics}. We discuss some of the issues in this section and provide suggestions on future directions. 
%\cll{\cite{corr/abs-1712-07107} and \cite{access/AkhtarM18} have discussed some issues in general way for computer vision applications.In this section, we will discuss the issues in following aspects: (1) perceivability; (2) semantics; (3) transferability;  (4) automation; and (5) unattacked textual neural networks 

\subsection{Perceivability} 
Perturbations in image pixels are usually hard to be perceived, thus do not affect human judgment, but can only fool the deep neural networks. 
However, the perturbation on text is obvious, no matter the perturbation is flipping  characters or changing words. Invalid  words and syntactic errors can be easily identified by human and detected by the grammar check software, hence the perturbation is hard to attack a real NLP system. 
However, many research works generate such types of adversarial examples. It is acceptable  only if the purpose is utilizing adversarial examples to robustify the attacked DNN models. 
In semantic-preserving perspective, changing a word in a sentence sometimes changes its semantics drastically and is easily detected by human beings. For NLP applications  such as reading comprehension, and  sentiment analysis, the adversarial examples need to be carefully designed in order not to change the \textit{should-be} output.  
%\ah{( I think there are something missing: in order not to change the \textit{should-be} output ,... what)} . 
Otherwise, both correct output and perturbed output change, \cll{violating} the purpose of generating adversarial examples. 
This is challenging and limited works reviewed considered this constraint. % \ah{(what these works: some references need)}.  
Therefore, for practical attack, we need to propose methods that make the perturbations not only unperceivable, but preserve correct grammar and semantics.

\subsection{Transferability}
Transferability is a common property for adversarial examples. It reflects the generalization of the attack methods. Transferability means adversarial examples generated for one deep neural network on a dataset can also effectively attack another deep neural network (i.e., cross-model generalization) or dataset (i.e., cross-data generalization).  This property is more often exploited in black-box attacks as the details of the deep neural networks does not affect the attack method \cll{much}. It is also shown that untargeted adversarial examples are much more transferable than targeted ones \cite{iclr/LiuCLS17}. 
Transferability can be \cll{organized into three levels in deep neural networks: (1) same architecture with different data; (2) different architectures with same application; (3) different architectures with different data~\cite{corr/abs-1712-07107}}.  Although current works on textual attacks cover both three levels, the performance of the transferred attacks still decrease drastically compared to it on the original architecture and data, i.e., poor generalization ability. More efforts are expected to deliver better generalization ability.

\subsection{Automation}
Some reviewed works are able to generate adversarial examples automatically, while others cannot. 
In white-box attacks, leveraging the loss function of the DNN can identify the most affected points (e.g., character, word) in a text automatically. Then the attacks are performed on these points by automatically modifying the corresponding texts. 
In black-box attacks, some attacks, e.g. \textit{substitution} train substitute DNNs and  apply white-box attack strategies on the substitution. This can be achieved automatically.  
However, most of the other  works craft the adversarial  examples in a manual manner. For example, \cite{emnlp/JiaL17} concatenated manually-chosen meaningless paragraphs to fool the reading comprehension systems, in order to discover the vulnerability of the victim DNNs.
Many research works followed their way, not aiming on practical attacks, \cll{but more on examining robustness of the target network}. 
\cll{These manaul works are time-consuming and impractical.  We believe that more efforts in this line could pass through this barrier in future.}

\subsection{New Architectures}
Although most of the common textual DNNs have gained attention from the \cll{perspective of adversarial attack} (Section \ref{sec:sub_texNN}), \cll{many} DNNs haven't been attacked so far. For example, the generative neural models: Generative Adversarial Networks (GANs)  and Variational Auto-Encoders  (VAEs). In  NLP,  they are used to generate texts. \cll{Deep generative models requires more sophisticated skill for model training. This would explain that these techniques have been mainly overlooked by adversarial attack so far}. Future works may consider about generating adversarial examples for these generative DNNs. 
Another example is differentiable neural computer (DNC). Only one work attacked DNC so far \cite{corr/ChanMXXLO2018}. 
Attention mechanism is somehow become a standard component in most of the sequential models. But there is no work examined the mechanism itself. Instead, works are either attack the overall system that contain attentions, or leverage  attention scores to identify the word for perturbation \cite{conll/BlohmJSYV18}.

\subsection{Iterative vs One-off}
Iterative attacks iteratively search and update the perturbations based on the gradient of the output of the attacked DNN model. Thus it shows high quality and effectiveness, that is the perturbations can be small enough and hard to defense. However, these methods usually require long time to find the proper perturbations, rendering an obstacle for attacking in real-time. Therefore, one-off attacks are proposed  to tackle this problem. FGSM \cite{iclr15/goodfellow} is one example of one-off attack. Natually, one-off attack is much faster than iterative attack, but is less effective and easier to be defensed \cite{corr/YuanHL2018}. 
%FSGM one off. L-GGBS iterative. 
When designing attack methods on a real application, attackers need to carefully consider the trade off between efficiency and effectiveness of the attack.

\section{Conclusion}
\label{sec:conclude}

This article presents the first comprehensive survey in the direction of generating textual adversarial examples on deep neural networks. We review recent research efforts and develop classification schemes to organize existing literature. Additionally we summarize and analyze them from different aspects. We attempt  to provide a good reference for researchers to gain insight of the challenges, methods and issues in this research topic and shed lights on future directions. 
We hope more robust deep neural models are proposed  based on the knowledge of the adversarial attacks. 

\bibliographystyle{ACM-Reference-Format}
\bibliography{nlp_adver}

\end{document}